\pdfoutput=1

\documentclass[11pt]{article}

\usepackage[table,xcdraw]{xcolor}

\usepackage[]{acl}

\usepackage{times}
\usepackage{latexsym}
\usepackage{graphicx}
\usepackage{cite}
\usepackage{multirow}
\usepackage{multicol}
\usepackage{soul}
\usepackage{footnotehyper}
\makesavenoteenv{tabular}
\usepackage{enumitem}
\usepackage{amsmath}

\usepackage[T1]{fontenc}

\usepackage[utf8]{inputenc}

\usepackage{microtype}

\usepackage{inconsolata}

\title{Authorship Obfuscation in Multilingual Machine-Generated Text Detection}

\author{Dominik Macko$^1$, Robert Moro$^1$, Adaku Uchendu$^2$, Ivan Srba$^1$, Jason Samuel Lucas$^3$, \\ {\bf Michiharu Yamashita}$^3$, {\bf Nafis Irtiza Tripto}$^3$, {\bf Dongwon Lee}$^3$,\\ 
{\bf Jakub Simko}$^1$, {\bf Maria Bielikova}$^1$ 
\vspace{0.05in} \\
  $^1$ Kempelen Institute of Intelligent Technologies, Slovakia\\
  \texttt{\{name.surname\}}@kinit.sk \\
  $^2$ MIT Lincoln Laboratory, USA\\
  \texttt{adaku.uchendu@ll.mit.edu} \\
  $^3$ The Pennsylvania State University, PA, USA \\
  \texttt{\{jsl5710, michiharu, nit5154, dongwon\}}@psu.edu \\}

\begin{document}
\maketitle

\begin{abstract}
High-quality text generation capability of recent Large Language Models (LLMs) causes concerns about their misuse (e.g., in massive generation/spread of disinformation). Machine-generated text (MGT) detection is important to cope with such threats. However, it is susceptible to authorship obfuscation (AO) methods, such as paraphrasing, which can cause MGTs to evade detection. So far, this was evaluated only in monolingual settings. Thus, the susceptibility of recently proposed multilingual detectors is still unknown. We fill this gap by comprehensively benchmarking the performance of 10 well-known AO methods, attacking 37 MGT detection methods
against MGTs in 11 languages (i.e., 10 $\times$ 37 $\times$ 11 = 4,070 combinations).
We also evaluate the effect of data augmentation on adversarial robustness using obfuscated texts. The results indicate that all tested AO methods can cause evasion of automated detection in all tested languages, where homoglyph attacks are especially successful. However, some of the AO methods severely damaged the text, making it no longer readable or easily recognizable by humans (e.g., changed language, weird characters).
\end{abstract}

\section{Introduction}
\label{sec:introduction}

Recent advances in Language Modeling have birthed Large Language Models (LLMs) which exhibit significant improvements, including the ability to generate texts easily misconstrued as human-written~\citep{zellers2019defending}. In addition, LLMs have been found to exhibit emergent abilities~\citep{wei2022emergent}, such as performing several NLP tasks at or over the human level, e.g., an LLM passing academic exams designed for humans in top 10\% of test takers \citep{openai2023gpt4}. While this is a celebratory feat for the NLP community, LLMs currently have several shortcomings like the generation of hate and biased speech \citep{deshpande2023toxicity,venkit2023nationality}, 
the generation of memorized and plagiarized content \citep{nasr2023scalable,carlini2021extracting,lee2023language}, and the generation of mis/disinformation \citep{lucas2023fighting, chen2023can, vykopal-etal-2024-disinformation, zellers2019defending}. 
Furthermore, we find that some of these LLMs like ChatGPT, Flan-T5 \citep{chung2022scaling}, Falcon \citep{almazrouei2023falcon}, LLaMA \citep{touvron2023llama}, etc. can have their alignment tuning bypassed by jailbreaking prompting techniques \citep{lucas2023fighting,chen2023combating}. 
Therefore, due to the great potential to be misused, it is imperative that at the minimum, we can accurately distinguish {\bf Machine-Generated Texts} ({\bf MGT}) from human-written ones. 


\begin{figure}
    \centering
    \includegraphics[width=\linewidth]{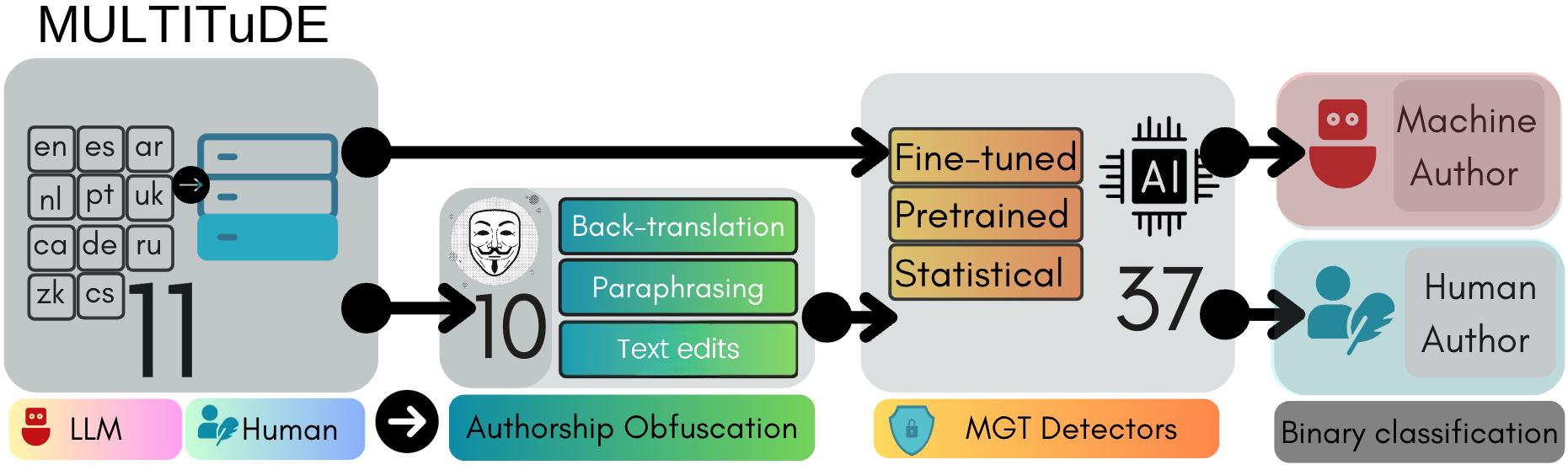}
    \caption{Benchmarking authorship obfuscation techniques for machine-generated text detection. 
    }
    \label{fig:framework}
    \vspace{-2mm}
\end{figure}

The MGT detection task has been studied very well for the binary case (\textit{Turing Test}), 
showing that accurate detection of machine-generated texts 
is possible \citep{uchendu2023attribution,crothers2023machine,wu2023survey}.
On the other hand, in this work, we investigate the less comprehensively studied problem
for MGT detection--\textbf{Authorship Obfuscation} ({\bf AO}), which refers to the process of masking an author's writing style/signature, while preserving semantics.
In the MGT detection task, AO represents an attack vector for MGT evading detection, being successful when true positives (i.e., detected MGT) are changed to false negatives (i.e., undetected MGT). 

We evaluate the robustness of MGT detectors to adversarial perturbations by implementing AO techniques (see \figurename~\ref{fig:framework} for an illustration). Since the task has been solely studied for monolingual settings, mostly English \citep{zellers2019defending,mitchell2023detectgpt}, but also Chinese or Russian \citep{Orzhenovskii2022DetectingAT,Pu2022UnravelingTM,guo-etal-2023-hc3}, we focus on the \textit{multilingual} perspective. Our key contributions are:

\textbf{(1)} We provide the first comprehensive \textbf{multilingual benchmark of AO methods to evade the MGT detection} by implementing 10 AO methods under 3 categories: \textit{Backtranslation}, \textit{Paraphrasing}, and \textit{Text edits}. We confirmed that eight of the ten used AO methods are usable in multilingual settings. Homoglyph-based attacks reached 70\% attack success in some languages.

\textbf{(2)} We provide \textbf{a unique dataset of multilingual obfuscated texts (human and machine-generated)}, which resulted into approximately 740k samples (although some of them contains the same text as original, i.e., the obfuscation failed). The dataset is available at Zenodo upon request for
research purposes only\footnote{\scriptsize\url{https://doi.org/10.5281/zenodo.13846588}}.

\textbf{(3)} We provide the first \textbf{evaluation of robustness of multilingual MGT detection methods} against authorship obfuscation. We evaluate 37 multilingual MGT detectors using the MULTITuDE \citep{macko2023multitude} multilingual benchmark dataset, containing 11 languages and 8 LLMs. Out of multiple novel findings, we found out, for example, that while basic \textit{ChatGPT paraphrasing} is not an effective AO technique, the \textit{homoglyph attacks} are very effective if not considered during detector training or text preprocessing.

\textbf{(4)} We evaluate the effect of \textbf{data augmentation using obfuscated texts on adversarial robustness} of multilingual detectors. We show that in most cases even simple data augmentation can improve the robustness of detectors to AO techniques.

\section{Related Work}
\label{sec:related}

With the advancements of LLMs, the MGT detection methods also adapt to multilingual settings \citep{wang2023m4, macko2023multitude}.
However, it is known that MGT detection is susceptible to AO methods (e.g. paraphrasing), which can drop the detection performance even by 90\% \citep{9892269, krishna2023paraphrasing, shi2023red}. As previously mentioned, AO methods have been evaluated in monolingual settings only. For some other NLP tasks, adversarial attacks (as a subset of AO) have been evaluated in multilingual settings \citep{rosenthal2021multilingual, wang-etal-2021-textflint}. However, multilingual evaluation of AO in MGT detection task is still missing, thus the severity of a potential threat of detection evasion is unknown.

\subsection{MGT Detection}
The current most popular MGT detectors are represented by two main groups: fine-tuned and statistical models. \textit{Fine-tuned models} are transformer-based models fine-tuned specifically for the MGT detection task \citep{Uchendu2021TURINGBENCHAB,Munir2021ThroughTL,Ai2022WhodunitLT,Liyanage2022ABC}. There are also available several models already \textit{pre-trained} for the MGT detection task, such as 
RoBERTa-base-OpenAI-Detector \citep{solaiman2019release}, Longformer detector \citep{li2023deepfake}, ChatGPT-Detector-RoBERTa-Chinese \citep{guo-etal-2023-hw}, GROVER detector \citep{zellers2019defending}, etc. These can be directly used in a zero-shot manner for detection of MGT. However, they are mostly monolingual. Multilingual models, such as XLM-RoBERTa \citep{DBLP:journals/corr/abs-1911-02116} or mBERT \citep{DBLP:journals/corr/abs-1810-04805}, can be fine-tuned for the task on a custom dataset to be used as multilingual detectors \citep{wang2023m4, macko2023multitude}.

\textit{Statistical models} use the statistical distribution of human-written and machine-generated texts to calculate the differences between the two. These detectors are usually based on a pre-trained LLM without fine-tuning, such as GPT-2 \citep{radford2019language}, mGPT \citep{https://doi.org/10.48550/arxiv.2204.07580}, 
Falcon \citep{almazrouei2023falcon}, etc.
to calculate a single or multiple metrics. Usually, a separate classifier (e.g., Logistic Regression or Random Forest) is trained to make a prediction based on such metrics. Popular statistical models include LLMDeviation \citep{wu2023mfd}, Rank \citep{gehrmann2019gltr}, GLTR Test 2 \citep{gehrmann2019gltr}, MFD \citep{wu2023mfd}, DetectGPT \citep{mitchell2023detectgpt}, or DetectLLM \citep{su2023detectllm}. There are new promising statistical methods available, such as Fast-DetectGPT \citep{bao2024fastdetectgpt}, 
GPT-who \citep{venkatraman2024gpt},
DNA-GPT \citep{yang2024dnagpt}, or Binoculars \citep{hans2024spotting}, which are however not covered in this study.

\subsection{MGT  Obfuscation}
To evaluate the adversarial robustness of MGT detectors, researchers have implemented several AO methods, which can be grouped into: 
\textit{Backtranslation}, 
\textit{Paraphrasing}, and
\textit{Text edits}.
\textit{Backtranslation} is the process of changing a text from one language to another and then back to the original (e.g., English $\to$ Spanish $\to$ English) \citep{10.1145/2660460.2660486,keswani2016author}. The idea is that the final output, the backtranslated version, will be subtly different from the original and thus, evade accurate detection. 
Backtranslation has been well studied in the authorship attribution niche field, making it suitable to apply to this task \citep{altakrori-etal-2022-multifaceted}. 
Next, \textit{Paraphrasing} is similar to the backtranslation method, however, the goal is to paraphrase/re-write the text and keep it in the same language as the original. It is currently the most popular AO technique in this detection task \citep{sadasivan2023can,krishna2023paraphrasing,lu2023large,shi2023red,koike2023outfox,tripto-etal-2024-ship}.
In \textit{Text edits}, the goal is to obfuscate authorship through one or more of the linguistic categories: Lexical, Syntactic, Morphological, or Othographic. Various text-edit attacks are available, such as 
lexical-based attacks implemented in GPTZzzs\footnote{\scriptsize\url{https://github.com/Declipsonator/GPTZzzs}} or DFTFooler \citep{pu2023deepfake}, a syntactic-based attack implemented in ALISON \citep{alison}, or an orthographic attack implemented in GPTZeroBypasser\footnote{\scriptsize\url{https://github.com/o2161405/GPTZero-Bypasser}}.

\section{Methodology}
\label{sec:methodology}

The methodology is illustrated in \figurename~\ref{fig:framework}, summarizing key components. 
We benchmark \ul{10 well-known AO methods attacking 37 MGT detection methods
against texts in 11 languages (i.e., 10 $\times$ 37 $\times$ 11 = 4,070 combinations).}
All source codes, data, and full results are publicly available to enable full replication of our study\footnote{\scriptsize\url{https://github.com/kinit-sk/mAO}}.

\subsection{Authorship Obfuscation Methods}
\label{sec:obfuscation}
To make the scope of this work feasible, we have used fully automated AO methods (leaving out human-in-the-loop methods) without any heavy modification (out-of-the-box, just changing the base model and/or tokenizer to a multilingual version if available). A single run of these methods have been used without any combinations (i.e., no iterative modification of the texts).
We have used 10 existing AO methods in this work (see Table~\ref{tab:ao_methods_overview}, links to and parameters of the AO methods are available in Appendix~\ref{sec:appendix-ao}), grouped into three main categories: backtranslation, paraphrasing, and text edits. We aimed to use at least two representatives of each group, while avoiding paid services and including at least one representative with naturally multilingual capability. 

\textbf{Backtranslation} uses English (as a high resource language) as an intermediary language for non-English texts and Spanish (as a representative from a different language family branch) for English texts. Since we have not found any dedicated multilingual \textbf{paraphrasing} method available, we have used ChatGPT instead (as used for this purpose in literature, e.g., \citealp{tripto-etal-2024-ship, cegin2023chatgpt}).
\textbf{Text edits} include synonym, homoglyph, as well as specialized adversarial attacks.


\begin{table}
\footnotesize 
\centering
\resizebox{\linewidth}{!}{
\begin{tabular}{m{0.2cm}|p{2.8cm}@{}p{6cm}}
\hline
& \textbf{AO Method} & \textbf{Description} \\
\hline
\multirow[c]{5}{*}{\rotatebox{90}{\scriptsize\textbf{\scriptsize{Backtranslation }} }} & \\
& m2m100\newline \citep{fan2020englishcentric} & A dedicated multi-language machine translation model.\\
& nllb-200\newline \citep{nllbteam2022language} & A research-purpose machine translation model.\\
\hline
\multirow[c]{8}{*}{\rotatebox{90}{\textbf{\scriptsize{Paraphrasing}}}} & Pegasus-paraphrase & A fine-tuned PEGASUS model for paraphrasing task, working on a sentence level.\\
& DIPPER\newline \citep{krishna2023paraphrasing} & A contextual paragraph-level paraphraser with a controllable diversity, based on English-only T5-xxl.\\
& ChatGPT & A popular OpenAI chat model. We have used the basic prompt of \texttt{``Paraphrase the following text in <language> language: <text>''}.\\
\hline
\multirow[c]{15}{*}{\rotatebox{90}{\textbf{\scriptsize{Text edits}}}} & GPTZzzs & A tool using an English dictionary of synonyms to replace a number of words.\\
& GPTZeroBypasser & A homoglyph attack to replace 9 specific Latin letters for Cyrillic letters and inserting a zero-width joiner pseudorandomly.\\
& HomoglyphAttack & Our generic version of a homoglyph attack using the whole confusables table to psudorandomly replace letters for their homoglyphs.\\
& ALISON\newline \citep{alison} & An adversarial perturbation attack that requires no queries to the target LLM, instead it targets its own classifier trained on most frequent ngrams from the train corpus.\\
& DFTFooler\newline \citep{pu2023deepfake} & An adversarial perturbation attack that attacks only machine-labeled samples, also requires no queries to the target model.\\
\hline
\end{tabular}
}
\caption{Authorship obfuscation methods overview.}
\label{tab:ao_methods_overview}
\vspace{-2mm}
\end{table}

\subsection{Dataset}
\label{sec:dataset}

As a source of original unobfuscated texts, we use the MULTITuDE benchmark dataset \citep{macko_2023_10013755} containing about 8k human-written news texts along with 66k texts generated by 8 LLMs (i.e., alpaca-lora, gpt-3.5-turbo, gpt-4, llama, opt, opt-iml-max, text-davinci-003, vicuna)
in 11 languages (i.e., Arabic, Catalan, Chinese, Czech, Dutch, English, German, Portuguese, Russian, Spanish, Ukrainian).
For each AO method, we duplicate the MULTITuDE data and replace the texts by their obfuscated version (we retry this if some obfuscated texts are the same as the unobfuscated ones; we limit the number of trials to 10).
We evaluate similarity between the obfuscated and the original texts using a range of similarity metrics along with a human check of data pseudo-random subset to gauge the quality of the obfuscation and detect potential pitfalls (see Appendix~\ref{sec:appendix-similarity}). 
\subsection{MGT Detection Methods}
\label{sec:detection}

The MGT detection methods used in this work are grouped into three categories (based on their nature): fine-tuned, pre-trained, and statistical.

We include \textbf{fine-tuned methods} which are based on multilingual language models (monolingual ones were omitted due to having low performance in the original MULTITuDE benchmark). They are trained on various portions of the train split of the MULTITuDE dataset (for monolingual and multilingual fine-tuning using all LLMs data) using the published scripts of the MULTITuDE benchmark\footnote{\scriptsize\url{https://github.com/kinit-sk/mgt-detection-benchmark}}. These include the following models: mDeBERTa-v3-base \citep{he2021debertav3},
XLM-RoBERTa-large \citep{DBLP:journals/corr/abs-1911-02116},
BERT-base-multilingual-cased \citep{DBLP:journals/corr/abs-1810-04805},
mGPT \citep{https://doi.org/10.48550/arxiv.2204.07580},
and OPT-IML-Max-1.3B \citep{iyer2022opt}.

As \textbf{pre-trained methods}, we include publicly available detection models that were fine-tuned for MGT detection tasks using different datasets (i.e., out-of-distribution data) and use them in a zero-shot manner in this work. These include RoBERTa-Base and 
RoBERTa-Large OpenAI Detectors 
\citep{solaiman2019release}, ChatGPT-detector-RoBERTa and 
ChatGPT-detector-RoBERTa-Chinese 
\citep{guo-etal-2023-hc3}, Longformer Detector 
\citep{li2023deepfake}, RoBERTa-base-autextification-Detection\footnote{\scriptsize\url{www.huggingface.co/arincon/roberta-base-autextification}}, and ruRoBERTa-ruatd-binary\footnote{\scriptsize\url{https://huggingface.co/orzhan/ruroberta-ruatd-binary}}.

As for \textbf{statistical methods}, we include \textit{single-metric methods} (Entropy by \citealp{10.5555/3053718.3053722}, LogLikelihood by \citealp{mitchell2023detectgpt}, Rank by \citealp{gehrmann2019gltr}, LogRank by \citealp{mitchell2023detectgpt}, LLMDeviation by \citealp{wu2023mfd}, DetectLLM-LRR by \citealp{su2023detectllm}), \textit{multi-metric methods} (GLTR Test 2 Features by \citealp{gehrmann2019gltr}, MFD by \citealp{wu2023mfd}), and \textit{perturbation-based methods} (DetectGPT by \citealp{mitchell2023detectgpt}, DetectLLM-NPR by \citealp{su2023detectllm}). Due to the long run-time of perturbation-based methods, these have been used only on a portion of the test set. All of the mentioned statistical methods are implemented in the publicly available IMGTB framework\footnote{\scriptsize\url{https://github.com/kinit-sk/IMGTB}} \citep{spiegel-macko-2024-imgtb}, which was used in its default configuration (i.e., Logistic Regression classifier for predictions without hyper-parameters optimization). Analogously to the fine-tuned methods, the classifiers of the statistical methods have been trained using the original MULTITuDE data only (i.e., unobfuscated).

\begin{table*}[!t]
\footnotesize 
\centering
\resizebox{\linewidth}{!}{
\begin{tabular}{clcccccc}
\hline
\bfseries Rank & \bfseries MGT Detection Method & \bfseries Category & \bfseries AUC ROC & \bfseries Macro avg. & \bfseries Macro avg. & \bfseries Macro avg. & \bfseries Macro avg. \\
& & & \bfseries (sorted) & \bfseries F1-score & \bfseries F1-score (optimal) & \bfseries F1-score (1\% FPR) & \bfseries F1-score (5\% FPR) \\
\hline
\bfseries 1 & XLM-RoBERTa-large (all) & F & 0.9247 & 0.5745 & 0.5119 & 0.2296 & 0.4538 \\
\bfseries 2 & XLM-RoBERTa-large (ru) & F & 0.9231 & 0.5983 & 0.4709 & 0.3298 & 0.4290 \\
\bfseries 3 & mDeBERTa-v3-base (all) & F & 0.9076 & 0.5388 & 0.4917 & 0.2826 & 0.4057 \\
\bfseries 4 & mDeBERTa-v3-base (ru) & F & 0.8895 & 0.6434 & 0.5160 & 0.1497 & 0.3405 \\
\bfseries 5 & mDeBERTa-v3-base (es) & F & 0.8616 & 0.5185 & 0.4743 & 0.1633 & 0.3573 \\
\bfseries 6 & BERT-base-multilingual-cased (all) & F & 0.8515 & 0.5215 & 0.4479 & 0.2176 & 0.3594 \\
\bfseries 7 & mGPT (all) & F & 0.8511 & 0.5347 & 0.4640 & 0.2525 & 0.3110 \\
\bfseries 8 & BERT-base-multilingual-cased (es) & F & 0.8505 & 0.5306 & 0.4764 & 0.0875 & 0.2648 \\
\bfseries 9 & mGPT (ru) & F & 0.8427 & 0.5901 & 0.4924 & 0.0110 & 0.0110 \\
\bfseries 10 & XLM-RoBERTa-large (es) & F & 0.8346 & 0.5035 & 0.4996 & 0.0110 & 0.2747 \\
\bfseries 11 & mGPT (es) & F & 0.8312 & 0.5074 & 0.4363 & 0.0110 & 0.3217 \\
\bfseries 12 & OPT-IML-Max-1.3B (all) & F & 0.8261 & 0.5265 & 0.4406 & 0.0110 & 0.1958 \\
\bfseries 13 & OPT-IML-Max-1.3B (es) & F & 0.7697 & 0.5024 & 0.4905 & 0.0110 & 0.0110 \\
\bfseries 14 & BERT-base-multilingual-cased (ru) & F & 0.7315 & 0.5072 & 0.3823 & 0.0602 & 0.0602 \\
\bfseries 15 & BERT-base-multilingual-cased (en) & F & 0.7198 & 0.4999 & 0.4361 & 0.1231 & 0.2303 \\
\bfseries 16 & OPT-IML-Max-1.3B (ru) & F & 0.7101 & 0.5346 & 0.4063 & 0.0776 & 0.1961 \\
\bfseries 17 & XLM-RoBERTa-large (en) & F & 0.6815 & 0.5285 & 0.3734 & 0.1274 & 0.2073 \\
\bfseries 18 & MFD & S & 0.6713 & 0.4799 & 0.3564 & 0.1069 & 0.2526 \\
\bfseries 19 & RoBERTa-large-OpenAI-Detector & P & 0.6618 & 0.2266 & 0.4972 & 0.4672 & 0.4997 \\
\bfseries 20 & Entropy & S & 0.6191 & 0.4972 & 0.2433 & 0.1562 & 0.2229 \\
\bfseries 21 & mGPT (en) & F & 0.6178 & 0.4936 & 0.3181 & 0.0110 & 0.0110 \\
\bfseries 22 & Longformer Detector & P & 0.6135 & 0.4972 & 0.2531 & 0.0582 & 0.1362 \\
\bfseries 23 & mDeBERTa-v3-base (en) & F & 0.6112 & 0.4660 & 0.4166 & 0.0152 & 0.0775 \\
\bfseries 24 & RoBERTa-base-OpenAI-Detector & P & 0.5955 & 0.1924 & 0.4972 & 0.4328 & 0.4897 \\
\bfseries 25 & OPT-IML-Max-1.3B (en) & F & 0.5824 & 0.5452 & 0.1489 & 0.0968 & 0.1489 \\
\bfseries 26 & DetectLLM-NPR & S & 0.5764 & 0.4926 & 0.2844 & 0.0636 & 0.1469 \\
\bfseries 27 & ChatGPT-Detector-RoBERTa-Chinese & P & 0.5585 & 0.3463 & 0.4107 & 0.0110 & 0.0110 \\
\bfseries 28 & GLTR Test 2 & S & 0.5385 & 0.4922 & 0.3001 & 0.0454 & 0.1188 \\
\bfseries 29 & DetectGPT & S & 0.5382 & 0.4926 & 0.2700 & 0.0581 & 0.1231 \\
\bfseries 30 & ChatGPT-Detector-RoBERTa & P & 0.5311 & 0.1036 & 0.4972 & 0.0110 & 0.4551 \\
\bfseries 31 & DetectLLM-LRR & S & 0.5250 & 0.4966 & 0.2587 & 0.0573 & 0.1494 \\
\bfseries 32 & RoBERTa-base-autextification-Detection & P & 0.4946 & 0.4883 & 0.0727 & 0.0727 & 0.0727 \\
\bfseries 33 & LogRank & S & 0.4669 & 0.4965 & 0.2504 & 0.0344 & 0.0635 \\
\bfseries 34 & LLMDeviation & S & 0.4589 & 0.4967 & 0.2349 & 0.0292 & 0.0664 \\
\bfseries 35 & LogLikelihood & S & 0.4508 & 0.4966 & 0.2521 & 0.0323 & 0.0595 \\
\bfseries 36 & ruRoBERTa-ruatd-binary & P & 0.4406 & 0.4772 & 0.0110 & 0.0110 & 0.0110 \\
\bfseries 37 & Rank & S & 0.3859 & 0.4972 & 0.0110 & 0.0110 & 0.0110 \\
\hline
\end{tabular}
}
\caption{Detection performance comparison of all MGT detection methods using all MULTITuDE test data (human and machine samples) and machine samples obfuscated by each AO method. \textit{Macro avg. F1-score (optimal)} is just a theoretical performance and refers to this metric after calibration of the classification threshold based on ROC curve maximizing difference between TPR (true positive rate) and FPR (false positive rate). Similarly, \textit{Macro avg. F1-score (1\% FPR) and (5\% FPR)} refer to this metric for the classification threshold closest to (but below) 1\% and 5\% of FPR. For the exact calculations of such thresholds, see the published source code. F refers to fine-tuned, P to pre-trained, and S to statistical MGT detection methods category.}
\label{tab:benchmark-all}
\end{table*}

\section{Experiments and Results}
\label{sec:results}

Firstly, our experiments focus on finding whether the existing AO methods (so far evaluated primarily for English) are even usable for non-English languages. Secondly, we focus on finding how robust the existing MGT detection methods are in regard to susceptibility to AO. Lastly, we evaluate the effect of using obfuscated texts for data augmentation to increase their adversarial robustness.

To compare MGT detection performance, we use standard metrics, namely \textbf{AUC ROC} (area under the curve of receiver operating characteristic) as a classification-threshold independent metric (not affected by a threshold calibration on a domain data) and \textbf{Macro avg. F1-score} as a metric balancing between a precision and a recall of the classification commonly used for the MGT detection task using imbalanced datasets. We further provide Macro avg. F1-score values theoretically achievable on the used data with the classification thresholds calibrated based on the ROC curve for optimal (i.e., maximal difference between TPR and FPR, where TPR is true positive rate and FPR is false positive rate), 1\% FPR and 5\% FPR conditions. We are aware that such a performance is not actually achievable due to calibration on the evaluation data (i.e., data leakage). Therefore, we are not using it to compare performance of the MGT detection methods, but to evaluate effect of AO on the detection (i.e., compare AO methods), while not being biased due to sub-optimal predictions.

We use two primary metrics to evaluate the effect of AO methods. \textbf{ASR} (attack success rate) measures the effectiveness of changing true positives
to false negatives,
considering all AO methods as attacks on detection. \textbf{AUC ROC drop} (detection performance decrease) is used for measuring the effect of obfuscated texts on detection performance.

Before diving deeper into the experiments, let us briefly examine the general detection performance on our data (i.e., difficulty of the data for this task). 
Table~\ref{tab:benchmark-all} shows a comparison of detection performance of all MGT detection methods using the whole test set (i.e., the original MULTITuDE human and machine texts from the test split and the obfuscated machine texts). Obfuscated human texts are not used in the experiments due to uncertain labeling after obfuscation (i.e., whether the texts modified by machine should still be considered as human) \citep{tripto-etal-2024-ship}. Based on the results of achieved AUC ROC values, existing detection methods can be also used for distinguishing between human-written and obfuscated machine-generated texts. However, based on Macro avg. F1-score maximum values, the data are quite challenging for the task. A high difference between AUC ROC and Macro avg. F1-score values are due to highly imbalanced data with a ratio between human and machine classes around 1:80.

\begin{table*}[!t]
\centering
\resizebox{0.95\linewidth}{!}{
\begin{tabular}{p{4.3cm}|p{1.2cm}p{1.2cm}p{1.2cm}p{1.2cm}p{1.2cm}p{1.2cm}p{1.2cm}p{1.2cm}p{1.2cm}p{1.2cm}p{1.2cm}|p{1.8cm}}
\hline
 & \multicolumn{11}{c|}{\bfseries Test Language [mean (±confidence interval)]} \\
\bfseries AO Method (Category) & \bfseries ar & \bfseries ca & \bfseries cs & \bfseries de & \bfseries en & \bfseries es & \bfseries nl & \bfseries pt & \bfseries ru & \bfseries uk & \bfseries zh & \bfseries $\rightarrow$ Average \\
\hline
\bfseries m2m100-1.2B (B) & {\cellcolor[HTML]{EBF7A3}} \textcolor{black}{0.1060 (±0.19)} & {\cellcolor[HTML]{C5E67E}} \textcolor{black}{0.2759 (±0.13)} & {\cellcolor[HTML]{E3F399}} \textcolor{black}{0.1429 (±0.16)} & {\cellcolor[HTML]{CDEA83}} \textcolor{black}{0.2451 (±0.13)} & {\cellcolor[HTML]{BDE379}} \textcolor{black}{0.3063 (±0.11)} & {\cellcolor[HTML]{D3EC87}} \textcolor{black}{0.2251 (±0.12)} & {\cellcolor[HTML]{C7E77F}} \textcolor{black}{0.2691 (±0.11)} & {\cellcolor[HTML]{CDEA83}} \textcolor{black}{0.2484 (±0.11)} & {\cellcolor[HTML]{D9EF8B}} \textcolor{black}{0.2014 (±0.14)} & {\cellcolor[HTML]{E3F399}} \textcolor{black}{0.1449 (±0.13)} & {\cellcolor[HTML]{DCF08F}} \textcolor{black}{0.1841 (±0.05)} & {\cellcolor[HTML]{D5ED88}} \textcolor{black}{0.2136} \\
\bfseries nllb-200-distilled-1.3B (B) & {\cellcolor[HTML]{ECF7A6}} \textcolor{black}{0.0983 (±0.01)} & {\cellcolor[HTML]{B9E176}} \textcolor{black}{0.3262 (±0.02)} & {\cellcolor[HTML]{D9EF8B}} \textcolor{black}{0.1980 (±0.01)} & {\cellcolor[HTML]{D9EF8B}} \textcolor{black}{0.1987 (±0.01)} & {\cellcolor[HTML]{D5ED88}} \textcolor{black}{0.2133 (±0.06)} & {\cellcolor[HTML]{DCF08F}} \textcolor{black}{0.1846 (±0.01)} & {\cellcolor[HTML]{D1EC86}} \textcolor{black}{0.2307 (±0.01)} & {\cellcolor[HTML]{D9EF8B}} \textcolor{black}{0.1990 (±0.01)} & {\cellcolor[HTML]{DCF08F}} \textcolor{black}{0.1868 (±0.01)} & {\cellcolor[HTML]{E2F397}} \textcolor{black}{0.1542 (±0.01)} & {\cellcolor[HTML]{C7E77F}} \textcolor{black}{0.2725 (±0.00)} & {\cellcolor[HTML]{D7EE8A}} \textcolor{black}{0.2057} \\
\bfseries Pegasus-paraphrase (P) & {\cellcolor[HTML]{DFF293}} \textcolor{black}{0.1706 (±0.03)} & {\cellcolor[HTML]{7DC765}} \textcolor{black}{0.5279 (±0.05)} & {\cellcolor[HTML]{A0D669}} \textcolor{black}{0.4143 (±0.04)} & {\cellcolor[HTML]{93D168}} \textcolor{black}{0.4564 (±0.08)} & {\cellcolor[HTML]{DAF08D}} \textcolor{black}{0.1912 (±0.08)} & {\cellcolor[HTML]{8ECF67}} \textcolor{black}{0.4730 (±0.06)} & {\cellcolor[HTML]{89CC67}} \textcolor{black}{0.4877 (±0.05)} & {\cellcolor[HTML]{69BE63}} \textcolor{black}{0.5883 (±0.05)} & {\cellcolor[HTML]{E5F49B}} \textcolor{black}{0.1395 (±0.05)} & {\cellcolor[HTML]{F5FBB2}} \textcolor{black}{0.0489 (±0.04)} & {\cellcolor[HTML]{C5E67E}} \textcolor{black}{0.2754 (±0.07)} & {\cellcolor[HTML]{B5DF74}} \textcolor{black}{0.3430} \\
\bfseries DIPPER (P) & {\cellcolor[HTML]{CDEA83}} \textcolor{black}{0.2425 (±0.03)} & {\cellcolor[HTML]{DCF08F}} \textcolor{black}{0.1830 (±0.06)} & {\cellcolor[HTML]{D7EE8A}} \textcolor{black}{0.2042 (±0.05)} & {\cellcolor[HTML]{D3EC87}} \textcolor{black}{0.2191 (±0.06)} & {\cellcolor[HTML]{E2F397}} \textcolor{black}{0.1556 (±0.05)} & {\cellcolor[HTML]{CDEA83}} \textcolor{black}{0.2478 (±0.05)} & {\cellcolor[HTML]{D5ED88}} \textcolor{black}{0.2120 (±0.04)} & {\cellcolor[HTML]{C9E881}} \textcolor{black}{0.2645 (±0.04)} & {\cellcolor[HTML]{E8F59F}} \textcolor{black}{0.1199 (±0.04)} & {\cellcolor[HTML]{F1F9AC}} \textcolor{black}{0.0719 (±0.05)} & {\cellcolor[HTML]{BFE47A}} \textcolor{black}{0.2991 (±0.12)} & {\cellcolor[HTML]{D9EF8B}} \textcolor{black}{0.2018} \\
\bfseries ChatGPT (P) & {\cellcolor[HTML]{EFF8AA}} \textcolor{black}{0.0803 (±0.00)} & {\cellcolor[HTML]{EBF7A3}} \textcolor{black}{0.1076 (±0.10)} & {\cellcolor[HTML]{ECF7A6}} \textcolor{black}{0.0958 (±0.06)} & {\cellcolor[HTML]{EBF7A3}} \textcolor{black}{0.1020 (±0.11)} & {\cellcolor[HTML]{E5F49B}} \textcolor{black}{0.1332 (±0.13)} & {\cellcolor[HTML]{EFF8AA}} \textcolor{black}{0.0798 (±0.11)} & {\cellcolor[HTML]{EEF8A8}} \textcolor{black}{0.0920 (±0.11)} & {\cellcolor[HTML]{F1F9AC}} \textcolor{black}{0.0766 (±0.12)} & {\cellcolor[HTML]{EBF7A3}} \textcolor{black}{0.1026 (±0.08)} & {\cellcolor[HTML]{EFF8AA}} \textcolor{black}{0.0831 (±0.03)} & {\cellcolor[HTML]{EBF7A3}} \textcolor{black}{0.1052 (±0.00)} & {\cellcolor[HTML]{ECF7A6}} \textcolor{black}{0.0962} \\
\bfseries GPTZzzs (T) & {\cellcolor[HTML]{FEFFBE}} \textcolor{black}{0.0053 (±0.14)} & {\cellcolor[HTML]{F1F9AC}} \textcolor{black}{0.0741 (±0.07)} & {\cellcolor[HTML]{FBFDBA}} \textcolor{black}{0.0173 (±0.11)} & {\cellcolor[HTML]{FAFDB8}} \textcolor{black}{0.0308 (±0.12)} & {\cellcolor[HTML]{A2D76A}} \textcolor{black}{0.4140 (±0.06)} & {\cellcolor[HTML]{EFF8AA}} \textcolor{black}{0.0849 (±0.13)} & {\cellcolor[HTML]{EFF8AA}} \textcolor{black}{0.0797 (±0.09)} & {\cellcolor[HTML]{EEF8A8}} \textcolor{black}{0.0928 (±0.13)} & {\cellcolor[HTML]{FEFFBE}} \textcolor{black}{0.0031 (±0.06)} & {\cellcolor[HTML]{FEFFBE}} \textcolor{black}{0.0010 (±0.05)} & {\cellcolor[HTML]{FDFEBC}} \textcolor{black}{0.0095 (±0.11)} & {\cellcolor[HTML]{F1F9AC}} \textcolor{black}{0.0739} \\
\bfseries GPTZeroBypasser (T) & {\cellcolor[HTML]{ABDB6D}} \textcolor{black}{0.3764 (±0.20)} & {\cellcolor[HTML]{8ECF67}} \textcolor{black}{0.4698 (±0.17)} & {\cellcolor[HTML]{D1EC86}} \textcolor{black}{0.2328 (±0.13)} & {\cellcolor[HTML]{96D268}} \textcolor{black}{0.4492 (±0.14)} & {\cellcolor[HTML]{6EC064}} \textcolor{black}{0.5752 (±0.17)} & {\cellcolor[HTML]{7AC665}} \textcolor{black}{0.5378 (±0.15)} & {\cellcolor[HTML]{70C164}} \textcolor{black}{0.5634 (±0.15)} & {\cellcolor[HTML]{6BBF64}} \textcolor{black}{0.5857 (±0.16)} & {\cellcolor[HTML]{8ECF67}} \textcolor{black}{0.4713 (±0.18)} & {\cellcolor[HTML]{AFDD70}} \textcolor{black}{0.3633 (±0.15)} & {\cellcolor[HTML]{DAF08D}} \textcolor{black}{0.1923 (±0.10)} & {\cellcolor[HTML]{98D368}} \textcolor{black}{0.4379} \\
\bfseries HomoglyphAttack (T) & {\cellcolor[HTML]{ABDB6D}} \textcolor{black}{0.3767 (±0.00)} & {\cellcolor[HTML]{39A758}} \textcolor{black}{0.7154 (±0.05)} & {\cellcolor[HTML]{93D168}} \textcolor{black}{0.4593 (±0.01)} & {\cellcolor[HTML]{54B45F}} \textcolor{black}{0.6440 (±0.02)} & {\cellcolor[HTML]{249D53}} \textcolor{black}{0.7684 (±0.14)} & {\cellcolor[HTML]{3CA959}} \textcolor{black}{0.7033 (±0.05)} & {\cellcolor[HTML]{2DA155}} \textcolor{black}{0.7447 (±0.04)} & {\cellcolor[HTML]{2DA155}} \textcolor{black}{0.7495 (±0.05)} & {\cellcolor[HTML]{7AC665}} \textcolor{black}{0.5371 (±0.00)} & {\cellcolor[HTML]{A0D669}} \textcolor{black}{0.4147 (±0.00)} & {\cellcolor[HTML]{E9F6A1}} \textcolor{black}{0.1131 (±0.00)} & {\cellcolor[HTML]{70C164}} \textcolor{black}{0.5660} \\
\bfseries ALISON (T) & {\cellcolor[HTML]{FBFDBA}} \textcolor{black}{0.0216 (±0.02)} & {\cellcolor[HTML]{F7FCB4}} \textcolor{black}{0.0451 (±0.04)} & {\cellcolor[HTML]{FBFDBA}} \textcolor{black}{0.0180 (±0.03)} & {\cellcolor[HTML]{FAFDB8}} \textcolor{black}{0.0275 (±0.03)} & {\cellcolor[HTML]{EBF7A3}} \textcolor{black}{0.1092 (±0.06)} & {\cellcolor[HTML]{FAFDB8}} \textcolor{black}{0.0312 (±0.02)} & {\cellcolor[HTML]{FAFDB8}} \textcolor{black}{0.0298 (±0.02)} & {\cellcolor[HTML]{F8FCB6}} \textcolor{black}{0.0340 (±0.02)} & {\cellcolor[HTML]{FBFDBA}} \textcolor{black}{0.0227 (±0.03)} & {\cellcolor[HTML]{FAFDB8}} \textcolor{black}{0.0261 (±0.03)} & {\cellcolor[HTML]{FDFEBC}} \textcolor{black}{0.0094 (±0.04)} & {\cellcolor[HTML]{F8FCB6}} \textcolor{black}{0.0340} \\
\bfseries DFTFooler (T) & {\cellcolor[HTML]{FEFFBE}} \textcolor{black}{0.0036 (±0.16)} & {\cellcolor[HTML]{D5ED88}} \textcolor{black}{0.2149 (±0.13)} & {\cellcolor[HTML]{E9F6A1}} \textcolor{black}{0.1105 (±0.16)} & {\cellcolor[HTML]{BDE379}} \textcolor{black}{0.3061 (±0.17)} & {\cellcolor[HTML]{BFE47A}} \textcolor{black}{0.3011 (±0.07)} & {\cellcolor[HTML]{D3EC87}} \textcolor{black}{0.2206 (±0.14)} & {\cellcolor[HTML]{BBE278}} \textcolor{black}{0.3147 (±0.14)} & {\cellcolor[HTML]{C7E77F}} \textcolor{black}{0.2678 (±0.15)} & {\cellcolor[HTML]{ECF7A6}} \textcolor{black}{0.0960 (±0.10)} & {\cellcolor[HTML]{F8FCB6}} \textcolor{black}{0.0372 (±0.05)} & {\cellcolor[HTML]{FDFEBC}} \textcolor{black}{0.0081 (±0.14)} & {\cellcolor[HTML]{DFF293}} \textcolor{black}{0.1710} \\
\hline
\bfseries $\downarrow$ Average & {\cellcolor[HTML]{E3F399}} \textcolor{black}{0.1481} & {\cellcolor[HTML]{C1E57B}} \textcolor{black}{0.2940} & {\cellcolor[HTML]{DAF08D}} \textcolor{black}{0.1893} & {\cellcolor[HTML]{C7E77F}} \textcolor{black}{0.2679} & {\cellcolor[HTML]{BBE278}} \textcolor{black}{0.3167} & {\cellcolor[HTML]{C5E67E}} \textcolor{black}{0.2788} & {\cellcolor[HTML]{BFE47A}} \textcolor{black}{0.3024} & {\cellcolor[HTML]{BDE379}} \textcolor{black}{0.3107} & {\cellcolor[HTML]{DAF08D}} \textcolor{black}{0.1880} & {\cellcolor[HTML]{E5F49B}} \textcolor{black}{0.1345} & {\cellcolor[HTML]{E3F399}} \textcolor{black}{0.1469} &  \\
\hline
\end{tabular}
}
\caption{Attack success rate of AO methods based on classification thresholds of individual MGT detection methods optimized for each language separately. Only human data and machine data generated by the selected best three LLM generators (gpt-4, gpt-3.5-turbo, vicuna) are used. Only
MGT detection methods achieving above 0.8 AUC ROC per each language on original test data are included (see Table~\ref{tab:benchmark-auc}). Per-language mean values of all detection methods are reported along with 95\% confidence interval error bounds. B refers to backtranslation, P to paraphrasing, and T to text edits AO method category.
 }
\label{tab:exp-asr}
\end{table*}

\subsection{Multilingual Capability of AO Methods}

In this experiment, we aim to answer the following research question: \textbf{RQ1:} \textit{Are the available out-of-the-box automated AO methods usable in multilingual settings?}
The objective is to find out whether an adversary can actually easily use the available methods to evade detection of MGTs in non-English languages. Is there a difference in the effect between languages (and their relationships)? Are there differences between AO method categories in transferability to non-English languages?

To answer these questions, we measure ASR of individual AO methods for each language by using optimal predictions of all MGT detection methods (i.e., the classification thresholds optimized to maximize the difference between TPR and FPR). Table~\ref{tab:exp-asr} shows the results, where a darker color represents a higher ASR value. For each test language, we test whether the differences between AO methods are statistically significant. To do this, we conduct repeated measures ANOVA tests for each test language: we use ASR for a given test language as a dependent variable, the MGT detection methods as ``subjects'' and the AO method as an independent within-subjects variable. For all 11 test languages, the observed differences are statistically significant ($p<0.05$). We further conduct post hoc pairwise tests between pairs of AO methods per each test language for a more in depth analysis.

To avoid biased results due to low-quality machine-generated texts, the results are reported for the data generated by the three best LLM generators (gpt-4, gpt-3.5-turbo, vicuna) based on the text similarity to human-written counterparts (see Table~\ref{tab:generatedtextanalysis}). Similarly, to avoid low-performing detectors affecting the mean values, we include only detection methods achieving above 0.8 AUC ROC per each test language on original test data (see Table~\ref{tab:benchmark-auc}). It results in averaging different sets of MGT detection methods for each language; therefore, we provide an ablation study including all detection methods and all LLM-generated data in Appendix~\ref{sec:appendix-ablation}.
The results might be also influenced by a potential overfit on the test data (due to calibration of the classification thresholds). Therefore, we also provide an ablation study with thresholds calibrated by using only 10\% of the original (unobfuscated) test data (it confirmed the conclusions).

\textbf{All (even English-only) AO methods are usable in multilingual settings.} The results indicate that each AO method is able to successfully cause evading detection at least for some samples. However, only in about 12\% of cases, the ASR reached 0.5 on average (representing about 50\% chance of the obfuscation being successful). Thus, a practical usage of most of AO methods is questionable.

\textbf{Effect of the English-only AO methods is inconsistent.} GPTZzzs affects mostly English, as expected since it uses an English dictionary of synonyms. Other English-only AO methods seem to have lower (the lowest in some cases) effect on English than on the other languages. Pegasus-paraphrase seems to achieve unusually high ASR on non-English languages, but not affecting non-Latin scripts as much. Similar behavior (regarding scripts) can be observed for DFTFooler, although in a lesser intensity. On the other hand, DIPPER seems to affect non-Latin scripts in a higher amount. A high ASR of Pegasus-parahrase cannot be solely due to changed language of the texts after obfuscation, since DIPPER changed the language for almost twice as many texts but achieves only about a half ASR.

\textbf{Homoglyph-based attacks are most successful across languages.} For most languages, these attacks offer above 50\% chance of successfully avoiding MGT detection. This effect can be expected since these attacks change the characters of the words, thus influencing inner representations (embeddings) in the detection models. However, such attacks can be easily detected/alleviated by preprocessing (e.g., by detecting multi-script texts). Out of the test languages, Chinese seems to be the most resistant to this kind of AO.

\textbf{Backtranslation and other paraphrasers are more successful than ChatGPT in obfusation.} However, we have used only a basic prompt for paraphrasing by ChatGPT. A more sophisticated prompt directing the writing style could help to increase the ASR. Nevertheless, the ChatGPT paraphrasing does not suffer by language change in multilingual settings as do the English-only dedicated paraphrasers (Pegasus-parphrase and DIPPER).

\begin{table*}[!t]
\centering
\resizebox{0.95\linewidth}{!}{
\begin{tabular}{m{0.5cm}m{0.5cm}|p{1.5cm}p{1.5cm}p{1.5cm}p{1.5cm}p{1.5cm}p{1.5cm}p{1.5cm}p{1.5cm}p{1.5cm}p{1.5cm}p{1.5cm}|p{2cm}}
\hline
& \bfseries  & \multicolumn{11}{c|}{\bfseries Test Language [mean (±confidence interval)]} \\
& \bfseries  & \bfseries ar & \bfseries ca & \bfseries cs & \bfseries de & \bfseries en & \bfseries es & \bfseries nl & \bfseries pt & \bfseries ru & \bfseries uk & \bfseries zh & \bfseries $\rightarrow$ Average \\
\hline
\multirow{8}{*}{\rotatebox{90}{\bfseries Train Language}} & \bfseries en & {\cellcolor[HTML]{FDD8A6}} \textcolor{black}{-12.73\% (±8.59\%)} & {\cellcolor[HTML]{FDC58E}} \textcolor{black}{-22.54\% (±8.37\%)} & {\cellcolor[HTML]{FDDDB1}} \textcolor{black}{-9.28\% (±6.10\%)} & {\cellcolor[HTML]{FDD9A8}} \textcolor{black}{-12.07\% (±5.41\%)} & {\cellcolor[HTML]{FEE6C4}} \textcolor{black}{-3.93\% (±3.00\%)} & {\cellcolor[HTML]{FDCC96}} \textcolor{black}{-19.20\% (±7.59\%)} & {\cellcolor[HTML]{FDC38C}} \textcolor{black}{-23.63\% (±8.75\%)} & {\cellcolor[HTML]{FDCA94}} \textcolor{black}{-19.79\% (±7.79\%)} & {\cellcolor[HTML]{FDD19B}} \textcolor{black}{-16.62\% (±7.83\%)} & {\cellcolor[HTML]{FDD5A0}} \textcolor{black}{-14.51\% (±7.44\%)} & {\cellcolor[HTML]{FEE0B8}} \textcolor{black}{-7.26\% (±2.97\%)} & {\cellcolor[HTML]{FDD49F}} \textcolor{black}{-14.69\%} \\
& \bfseries es & {\cellcolor[HTML]{FEE4BF}} \textcolor{black}{-5.28\% (±4.99\%)} & {\cellcolor[HTML]{FEDFB5}} \textcolor{black}{-8.21\% (±5.30\%)} & {\cellcolor[HTML]{FEE7C5}} \textcolor{black}{-3.64\% (±4.02\%)} & {\cellcolor[HTML]{FEE0B7}} \textcolor{black}{-7.67\% (±4.01\%)} & {\cellcolor[HTML]{FDC791}} \textcolor{black}{-21.54\% (±10.07\%)} & {\cellcolor[HTML]{FEE4C0}} \textcolor{black}{-5.15\% (±3.60\%)} & {\cellcolor[HTML]{FDDDB1}} \textcolor{black}{-9.31\% (±5.50\%)} & {\cellcolor[HTML]{FEDFB5}} \textcolor{black}{-8.28\% (±5.39\%)} & {\cellcolor[HTML]{FDDAAB}} \textcolor{black}{-11.24\% (±5.95\%)} & {\cellcolor[HTML]{FEDFB5}} \textcolor{black}{-8.02\% (±5.00\%)} & {\cellcolor[HTML]{FEE8C8}} \textcolor{black}{-2.86\% (±1.75\%)} & {\cellcolor[HTML]{FEDFB5}} \textcolor{black}{-8.29\%} \\
& \bfseries ru & {\cellcolor[HTML]{FEE6C4}} \textcolor{black}{-3.90\% (±3.69\%)} & {\cellcolor[HTML]{FDD8A7}} \textcolor{black}{-12.58\% (±6.61\%)} & {\cellcolor[HTML]{FEE7C7}} \textcolor{black}{-3.22\% (±2.99\%)} & {\cellcolor[HTML]{FEE5C1}} \textcolor{black}{-4.70\% (±4.67\%)} & {\cellcolor[HTML]{FDD09A}} \textcolor{black}{-17.11\% (±9.37\%)} & {\cellcolor[HTML]{FEE2BB}} \textcolor{black}{-6.68\% (±4.94\%)} & {\cellcolor[HTML]{FDD9AA}} \textcolor{black}{-11.50\% (±5.51\%)} & {\cellcolor[HTML]{FEDFB5}} \textcolor{black}{-8.03\% (±5.99\%)} & {\cellcolor[HTML]{FEE8C8}} \textcolor{black}{-2.80\% (±1.50\%)} & {\cellcolor[HTML]{FEE7C5}} \textcolor{black}{-3.30\% (±1.94\%)} & {\cellcolor[HTML]{FEE6C4}} \textcolor{black}{-3.87\% (±1.87\%)} & {\cellcolor[HTML]{FEE1B9}} \textcolor{black}{-7.06\%} \\
\cline{2-14}
& \bfseries all & {\cellcolor[HTML]{FEE5C1}} \textcolor{black}{-4.77\% (±4.05\%)} & {\cellcolor[HTML]{FEDFB4}} \textcolor{black}{-8.72\% (±4.27\%)} & {\cellcolor[HTML]{FEE6C4}} \textcolor{black}{-4.01\% (±2.60\%)} & {\cellcolor[HTML]{FEDFB5}} \textcolor{black}{-8.00\% (±3.75\%)} & {\cellcolor[HTML]{FEE8C9}} \textcolor{black}{-2.14\% (±1.79\%)} & {\cellcolor[HTML]{FEE5C1}} \textcolor{black}{-4.65\% (±2.68\%)} & {\cellcolor[HTML]{FDDAAB}} \textcolor{black}{-11.11\% (±4.73\%)} & {\cellcolor[HTML]{FEE2BB}} \textcolor{black}{-6.71\% (±3.59\%)} & {\cellcolor[HTML]{FEE5C3}} \textcolor{black}{-4.35\% (±2.23\%)} & {\cellcolor[HTML]{FEE5C1}} \textcolor{black}{-4.56\% (±2.59\%)} & {\cellcolor[HTML]{FEE5C3}} \textcolor{black}{-4.27\% (±2.09\%)} & {\cellcolor[HTML]{FEE3BD}} \textcolor{black}{-5.75\%} \\
\hline
\hline
\multirow{6}{*}{\rotatebox{90}{\bfseries Category}} & \bfseries F & {\cellcolor[HTML]{FEE2BB}} \textcolor{black}{-6.67\% (±2.80\%)} & {\cellcolor[HTML]{FDD8A6}} \textcolor{black}{-13.01\% (±3.19\%)} & {\cellcolor[HTML]{FEE4C0}} \textcolor{black}{-5.04\% (±2.05\%)} & {\cellcolor[HTML]{FEDFB5}} \textcolor{black}{-8.11\% (±2.22\%)} & {\cellcolor[HTML]{FDDAAB}} \textcolor{black}{-11.18\% (±3.67\%)} & {\cellcolor[HTML]{FDDEB3}} \textcolor{black}{-8.92\% (±2.62\%)} & {\cellcolor[HTML]{FDD6A2}} \textcolor{black}{-13.89\% (±3.18\%)} & {\cellcolor[HTML]{FDDBAD}} \textcolor{black}{-10.70\% (±2.96\%)} & {\cellcolor[HTML]{FDDEB3}} \textcolor{black}{-8.75\% (±2.61\%)} & {\cellcolor[HTML]{FEE0B7}} \textcolor{black}{-7.60\% (±2.40\%)} & {\cellcolor[HTML]{FEE5C1}} \textcolor{black}{-4.57\% (±1.11\%)} & {\cellcolor[HTML]{FDDEB3}} \textcolor{black}{-8.95\%} \\
& \bfseries P & {\cellcolor[HTML]{FDDDB0}} \textcolor{black}{-9.58\% (±7.23\%)} & {\cellcolor[HTML]{FEE7C7}} \textcolor{black}{-3.07\% (±6.73\%)} & {\cellcolor[HTML]{FEE3BD}} \textcolor{black}{-5.94\% (±6.22\%)} & {\cellcolor[HTML]{FEE2BB}} \textcolor{black}{-6.76\% (±6.72\%)} & {\cellcolor[HTML]{FDCA93}} \textcolor{black}{-20.46\% (±7.89\%)} & {\cellcolor[HTML]{FEE1B9}} \textcolor{black}{-7.00\% (±6.71\%)} & {\cellcolor[HTML]{FEE7C5}} \textcolor{black}{-3.63\% (±6.04\%)} & {\cellcolor[HTML]{FEDFB5}} \textcolor{black}{-8.30\% (±6.26\%)} & {\cellcolor[HTML]{FEE6C4}} \textcolor{black}{-3.91\% (±7.86\%)} & {\cellcolor[HTML]{FEE3BD}} \textcolor{black}{-5.73\% (±6.53\%)} & {\cellcolor[HTML]{FEEACE}} \textcolor{black}{-0.59\% (±4.98\%)} & {\cellcolor[HTML]{FEE1B9}} \textcolor{black}{-6.82\%} \\
& \bfseries S & {\cellcolor[HTML]{FEE2BC}} \textcolor{black}{-6.22\% (±7.36\%)} & {\cellcolor[HTML]{FDDDB1}} \textcolor{black}{-9.28\% (±10.31\%)} & {\cellcolor[HTML]{FEE1B9}} \textcolor{black}{-7.10\% (±9.28\%)} & {\cellcolor[HTML]{FDD6A3}} \textcolor{black}{-13.66\% (±8.20\%)} & {\cellcolor[HTML]{FDCE98}} \textcolor{black}{-17.79\% (±9.98\%)} & {\cellcolor[HTML]{FDD5A0}} \textcolor{black}{-14.32\% (±8.14\%)} & {\cellcolor[HTML]{FDDCAF}} \textcolor{black}{-10.04\% (±8.69\%)} & {\cellcolor[HTML]{FDD8A6}} \textcolor{black}{-12.76\% (±8.20\%)} & {\cellcolor[HTML]{FDCB95}} \textcolor{black}{-19.42\% (±7.95\%)} & {\cellcolor[HTML]{FDD5A0}} \textcolor{black}{-14.54\% (±8.54\%)} & {\cellcolor[HTML]{FEE2BC}} \textcolor{black}{-6.22\% (±3.50\%)} & {\cellcolor[HTML]{FDD9A8}} \textcolor{black}{-11.94\%} \\
\hline
\end{tabular}
}
\caption{AUC ROC drop of differently fine-tuned MGT detection methods (top) and of different categories (F -- fine-tuned, P -- pre-trained, S -- statistical) of MGT detection methods (bottom). Per-language mean value is reported for each group (train language or category) along with 95\% confidence interval error bounds. Only test data of AO methods passing the quality check are used.}
\label{tab:exp-drop}
\end{table*}

\subsubsection{Human Validation of Readability}
Such an experiment properly evaluates the impact of AO methods on the automated MGT detection methods. However, the AO methods could have severely damaged the texts, which would be either unreadable by humans or clearly suspicious. Therefore, we have done a manual human check (see Appendix~\ref{sec:appendix-similarity}) of the obfuscated data sample by three human evaluators (average Pearson correlation between annotations of 0.7, total annotation agreement accuracy of 0.65 with majority 2 of 3 accuracy of 0.99). Based on such a quality check along with text-similarity analysis between unobfuscated and obfuscated texts, Pegasus-paraphrase and DIPPER AO methods are unusable in the multilingual settings. Therefore, these two methods are disqualified from the further experiments (although included in the Appendices for a reference).

\subsection{Adversarial Robustness of Multilingual MGT Detection Methods}

This experiment aims to answer the following research question: \textbf{RQ2:} \textit{How robust are multilingual MGT detection methods against out-of-the-box automated AO methods?} Namely, is there a difference in robustness between languages or MGT detection method categories? Also, is there a difference between monolingually and multilingually fine-tuned MGT detection methods?

To answer these questions, we measure general detection performance (AUC ROC) drop caused by texts being obfuscated by individual AO methods for each language separately. A smaller AUC ROC drop (a higher value) means a detection method is more robust against obfuscation.
To see differences, we provide aggregated results based on train language used for fine-tuning and based on MGT detection method category in Table~\ref{tab:exp-drop}. Analogously to the previous experiment, we have tested statistical significance for each test language. AUC ROC and AUC ROC drop values per each AO method and each test language is reported in Appendix~\ref{sec:appendix-results}.

\textbf{There are clear differences among languages and among MGT detection method categories.}
All of them are most sensitive to homoglyph attacks, both of the attacks having similar effect on fine-tuned and pre-trained methods. However, statistical methods are more resistant to GPTZeroBypasser and more vulnerable to HomoglyphAttack. Fine-tuned and statistical methods are also affected by backtranslation, unlike pre-trained methods. 

\textbf{English is most vulnerable to obfuscation in pre-trained methods.} However, this might be influenced by being mostly represented by English-only detection models. In fine-tuned and statistical methods, Arabic, Czech and Chinese are more resistant to obfuscation than the others. However, these observations are not statistically significant due to high variation between AO methods.

\textbf{English-only fine-tuned MGT detection methods are least robust against obfuscation.} These methods are affected more by obfuscation of non-English texts. On the other hand, detection methods fine-tuned monolingually by Spanish or Russian are mostly sensitive to English texts obfuscation. Multilingually fine-tuned detection methods are the most robust, although difference between them and the Russian or Spanish monolingually fine-tuned methods are not statistically significant. 

\subsection{Multilingual Obfuscation for Data Augmentation}

Finally, this experiment aims to answer the following research question: \textbf{RQ3:} \textit{Does the training of multilingual MGT detection methods also on obfuscated data increase their adversarial robustness for individual languages?} Is the effect the same when using texts obfuscated by individual AO methods separately and by all of them combined?

To answer these questions, we measure AUC ROC and focus on a difference before and after adversarial retraining by data augmentation. We refer to these detection methods as originally trained and adversarially trained. For this experiment, we use only multilingually fine-tuned detection methods to limit the scope. As in the previous experiments, only obfuscated machine-class samples (of AO methods passing the quality check) were used for augmenting the train data for adversarial retraining. This caused highly imbalanced (human vs. machine classes) train set; therefore, we have pseudorandomly upsampled the minority class using a simple duplication method.
For the results not to be biased due to this difference in the fine-tuning process, we have also retrained original detection methods in this manner. For individual AO methods data adversarial retraining, we have selected one from backtranslation, one from paraphrasing, one from homoglyph attacks, and one from adversarial attacks, based on the highest absolute number of successfully obfuscated texts. For each adversarial retraining process, we used the same amounts of original and obfuscated samples.

\begin{figure}[!b]
    \centering
    \includegraphics[width=\linewidth]{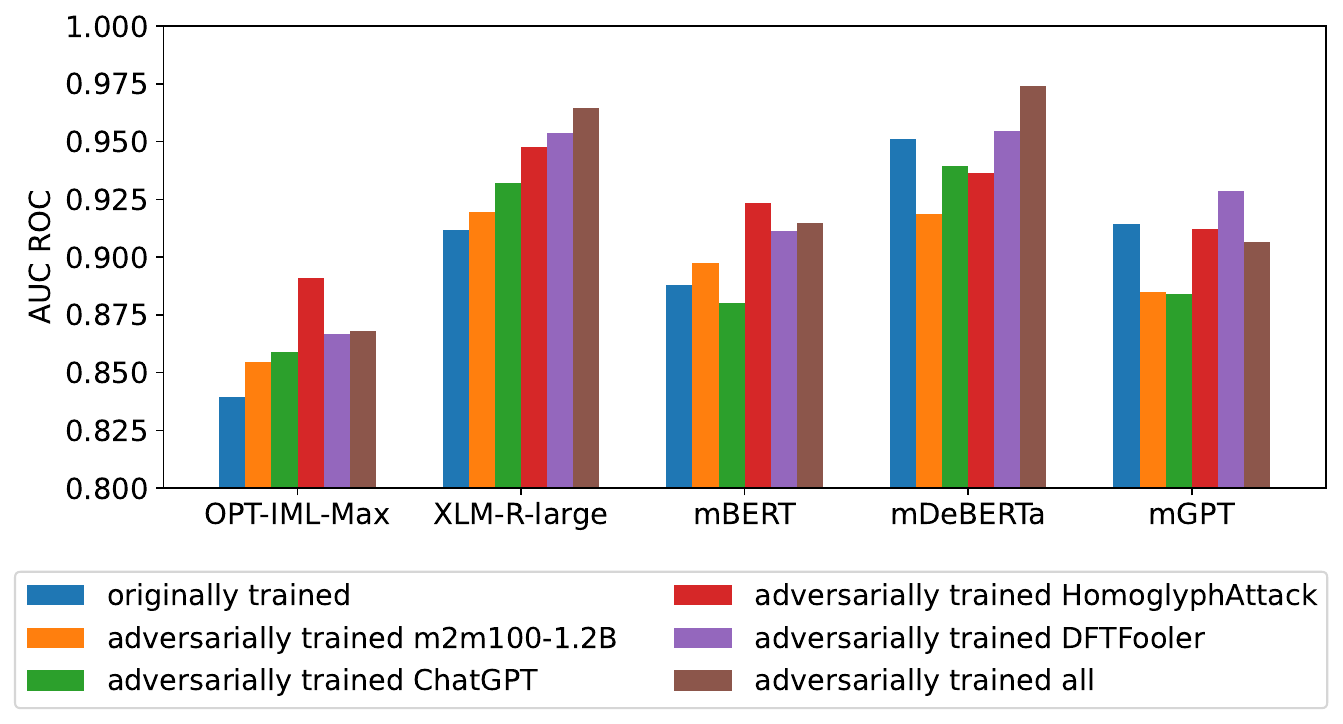}
    \caption{Detection performance (AUC ROC) of originally (leftmost bar) and adversarially trained MGT detection methods on the test data including all unobfuscated texts and obfuscated texts by AO methods passing the quality check. Note that the y-axis starts at 0.8.
    }
    \label{fig:fig-at}
\end{figure}
\begin{table*}[!t]
\footnotesize 

\centering
\resizebox{\linewidth}{!}{
\begin{tabular}{l|c|ccccc||ccccc}
\hline
 & \multicolumn{6}{c||}{\bfseries Absolute AUC ROC $\uparrow$} & \multicolumn{5}{c}{\bfseries Relative AUC ROC Difference $\downarrow$} \\
\bfseries AO Method (Category) & \rotatebox[origin=l]{90}{\bfseries \parbox{1.5cm}{originally trained}} & \rotatebox[origin=l]{90}{\bfseries all} & \rotatebox[origin=l]{90}{\bfseries \parbox{1.5cm}{m2m100-1.2B}} & \rotatebox[origin=l]{90}{\bfseries \parbox{1.5cm}{ChatGPT}} & \rotatebox[origin=l]{90}{\bfseries \parbox{1.5cm}{Homoglyph Attack}} & \rotatebox[origin=l]{90}{\bfseries \parbox{1.5cm}{DFTFooler}} & \rotatebox[origin=l]{90}{\bfseries all} & \rotatebox[origin=l]{90}{\bfseries \parbox{1.5cm}{m2m100-1.2B}} & \rotatebox[origin=l]{90}{\bfseries \parbox{1.5cm}{ChatGPT}} & \rotatebox[origin=l]{90}{\bfseries \parbox{1.5cm}{Homoglyph Attack}} & \rotatebox[origin=l]{90}{\bfseries \parbox{1.5cm}{DFTFooler}} \\
\hline
\bfseries original & {\cellcolor[HTML]{B3C3DE}} \color[HTML]{000000} 0.9372 & {\cellcolor[HTML]{BDC8E1}} \color[HTML]{000000} 0.9139 & {\cellcolor[HTML]{B7C5DF}} \color[HTML]{000000} 0.9275 & {\cellcolor[HTML]{B5C4DF}} \color[HTML]{000000} 0.9312 & {\cellcolor[HTML]{B7C5DF}} \color[HTML]{000000} 0.9270 & {\cellcolor[HTML]{B5C4DF}} \color[HTML]{000000} 0.9317 & {\cellcolor[HTML]{FFFAB6}} \color[HTML]{000000} -2.54\% & {\cellcolor[HTML]{FFFDBC}} \color[HTML]{000000} -1.07\% & {\cellcolor[HTML]{FFFDBC}} \color[HTML]{000000} -0.64\% & {\cellcolor[HTML]{FFFDBC}} \color[HTML]{000000} -1.09\% & {\cellcolor[HTML]{FFFDBC}} \color[HTML]{000000} -0.58\% \\
\bfseries m2m100-1.2B (B) & {\cellcolor[HTML]{C1CAE2}} \color[HTML]{000000} 0.9069 & {\cellcolor[HTML]{C5CCE3}} \color[HTML]{000000} 0.8985 & {\cellcolor[HTML]{B1C2DE}} \color[HTML]{000000} 0.9392 & {\cellcolor[HTML]{C4CBE3}} \color[HTML]{000000} 0.9019 & {\cellcolor[HTML]{C9CEE4}} \color[HTML]{000000} 0.8911 & {\cellcolor[HTML]{C6CCE3}} \color[HTML]{000000} 0.8951 & {\cellcolor[HTML]{FFFDBC}} \color[HTML]{000000} -1.02\% & {\cellcolor[HTML]{F5FBB2}} \color[HTML]{000000} 3.56\% & {\cellcolor[HTML]{FFFDBC}} \color[HTML]{000000} -0.57\% & {\cellcolor[HTML]{FFFBB8}} \color[HTML]{000000} -1.80\% & {\cellcolor[HTML]{FFFCBA}} \color[HTML]{000000} -1.32\% \\
\bfseries nllb-200-distilled-1.3B (B) & {\cellcolor[HTML]{C1CAE2}} \color[HTML]{000000} 0.9060 & {\cellcolor[HTML]{C5CCE3}} \color[HTML]{000000} 0.8989 & {\cellcolor[HTML]{B9C6E0}} \color[HTML]{000000} 0.9234 & {\cellcolor[HTML]{BBC7E0}} \color[HTML]{000000} 0.9214 & {\cellcolor[HTML]{C9CEE4}} \color[HTML]{000000} 0.8900 & {\cellcolor[HTML]{C6CCE3}} \color[HTML]{000000} 0.8957 & {\cellcolor[HTML]{FFFDBC}} \color[HTML]{000000} -0.87\% & {\cellcolor[HTML]{FAFDB8}} \color[HTML]{000000} 1.89\% & {\cellcolor[HTML]{FAFDB8}} \color[HTML]{000000} 1.70\% & {\cellcolor[HTML]{FFFBB8}} \color[HTML]{000000} -1.81\% & {\cellcolor[HTML]{FFFCBA}} \color[HTML]{000000} -1.14\% \\
\bfseries ChatGPT (P) & {\cellcolor[HTML]{B8C6E0}} \color[HTML]{000000} 0.9254 & {\cellcolor[HTML]{BCC7E1}} \color[HTML]{000000} 0.9169 & {\cellcolor[HTML]{B8C6E0}} \color[HTML]{000000} 0.9258 & {\cellcolor[HTML]{A8BEDC}} \color[HTML]{000000} 0.9587 & {\cellcolor[HTML]{BFC9E1}} \color[HTML]{000000} 0.9132 & {\cellcolor[HTML]{BFC9E1}} \color[HTML]{000000} 0.9116 & {\cellcolor[HTML]{FFFDBC}} \color[HTML]{000000} -0.97\% & {\cellcolor[HTML]{FEFFBE}} \color[HTML]{000000} 0.01\% & {\cellcolor[HTML]{F5FBB2}} \color[HTML]{000000} 3.67\% & {\cellcolor[HTML]{FFFCBA}} \color[HTML]{000000} -1.35\% & {\cellcolor[HTML]{FFFCBA}} \color[HTML]{000000} -1.52\% \\
\bfseries GPTZzzs (T) & {\cellcolor[HTML]{B5C4DF}} \color[HTML]{000000} 0.9311 & {\cellcolor[HTML]{BCC7E1}} \color[HTML]{000000} 0.9170 & {\cellcolor[HTML]{B9C6E0}} \color[HTML]{000000} 0.9216 & {\cellcolor[HTML]{B8C6E0}} \color[HTML]{000000} 0.9258 & {\cellcolor[HTML]{B9C6E0}} \color[HTML]{000000} 0.9221 & {\cellcolor[HTML]{B4C4DF}} \color[HTML]{000000} 0.9349 & {\cellcolor[HTML]{FFFCBA}} \color[HTML]{000000} -1.58\% & {\cellcolor[HTML]{FFFDBC}} \color[HTML]{000000} -1.05\% & {\cellcolor[HTML]{FFFDBC}} \color[HTML]{000000} -0.57\% & {\cellcolor[HTML]{FFFDBC}} \color[HTML]{000000} -0.97\% & {\cellcolor[HTML]{FEFFBE}} \color[HTML]{000000} 0.41\% \\
\bfseries GPTZeroBypasser (T) & {\cellcolor[HTML]{DAD9EA}} \color[HTML]{000000} 0.8443 & {\cellcolor[HTML]{9EBAD9}} \color[HTML]{000000} 0.9783 & {\cellcolor[HTML]{DEDCEC}} \color[HTML]{000000} 0.8316 & {\cellcolor[HTML]{D7D6E9}} \color[HTML]{000000} 0.8548 & {\cellcolor[HTML]{A9BFDC}} \color[HTML]{000000} 0.9554 & {\cellcolor[HTML]{BBC7E0}} \color[HTML]{000000} 0.9197 & {\cellcolor[HTML]{D1EC86}} \color[HTML]{000000} 16.27\% & {\cellcolor[HTML]{FFFBB8}} \color[HTML]{000000} -1.71\% & {\cellcolor[HTML]{FBFDBA}} \color[HTML]{000000} 1.37\% & {\cellcolor[HTML]{DAF08D}} \color[HTML]{000000} 13.43\% & {\cellcolor[HTML]{E6F59D}} \color[HTML]{000000} 9.07\% \\
\bfseries HomoglyphAttack (T) & {\cellcolor[HTML]{D6D6E9}} \color[HTML]{000000} 0.8580 & {\cellcolor[HTML]{9FBAD9}} \color[HTML]{000000} 0.9760 & {\cellcolor[HTML]{E1DFED}} \color[HTML]{000000} 0.8219 & {\cellcolor[HTML]{E2DFEE}} \color[HTML]{000000} 0.8183 & {\cellcolor[HTML]{96B6D7}} \color[HTML]{000000} 0.9903 & {\cellcolor[HTML]{AFC1DD}} \color[HTML]{000000} 0.9453 & {\cellcolor[HTML]{D9EF8B}} \color[HTML]{000000} 14.15\% & {\cellcolor[HTML]{FFF6B0}} \color[HTML]{000000} -4.34\% & {\cellcolor[HTML]{FFF5AE}} \color[HTML]{000000} -4.63\% & {\cellcolor[HTML]{D1EC86}} \color[HTML]{000000} 15.91\% & {\cellcolor[HTML]{E2F397}} \color[HTML]{000000} 10.40\% \\
\bfseries ALISON (T) & {\cellcolor[HTML]{B4C4DF}} \color[HTML]{000000} 0.9328 & {\cellcolor[HTML]{B4C4DF}} \color[HTML]{000000} 0.9346 & {\cellcolor[HTML]{B8C6E0}} \color[HTML]{000000} 0.9244 & {\cellcolor[HTML]{B8C6E0}} \color[HTML]{000000} 0.9252 & {\cellcolor[HTML]{BBC7E0}} \color[HTML]{000000} 0.9215 & {\cellcolor[HTML]{B9C6E0}} \color[HTML]{000000} 0.9234 & {\cellcolor[HTML]{FEFFBE}} \color[HTML]{000000} 0.16\% & {\cellcolor[HTML]{FFFDBC}} \color[HTML]{000000} -0.93\% & {\cellcolor[HTML]{FFFDBC}} \color[HTML]{000000} -0.80\% & {\cellcolor[HTML]{FFFCBA}} \color[HTML]{000000} -1.22\% & {\cellcolor[HTML]{FFFDBC}} \color[HTML]{000000} -0.99\% \\
\bfseries DFTFooler (T) & {\cellcolor[HTML]{BCC7E1}} \color[HTML]{000000} 0.9172 & {\cellcolor[HTML]{B5C4DF}} \color[HTML]{000000} 0.9306 & {\cellcolor[HTML]{C2CBE2}} \color[HTML]{000000} 0.9031 & {\cellcolor[HTML]{C1CAE2}} \color[HTML]{000000} 0.9074 & {\cellcolor[HTML]{BCC7E1}} \color[HTML]{000000} 0.9183 & {\cellcolor[HTML]{A5BDDB}} \color[HTML]{000000} 0.9626 & {\cellcolor[HTML]{FBFDBA}} \color[HTML]{000000} 1.44\% & {\cellcolor[HTML]{FFFCBA}} \color[HTML]{000000} -1.58\% & {\cellcolor[HTML]{FFFDBC}} \color[HTML]{000000} -1.05\% & {\cellcolor[HTML]{FEFFBE}} \color[HTML]{000000} 0.14\% & {\cellcolor[HTML]{F1F9AC}} \color[HTML]{000000} 5.04\% \\
\hline
\bfseries $\downarrow$ Average & {\cellcolor[HTML]{C1CAE2}} \color[HTML]{000000} 0.9065 & {\cellcolor[HTML]{B7C5DF}} \color[HTML]{000000} 0.9294 & {\cellcolor[HTML]{C4CBE3}} \color[HTML]{000000} 0.9021 & {\cellcolor[HTML]{C2CBE2}} \color[HTML]{000000} 0.9050 & {\cellcolor[HTML]{B8C6E0}} \color[HTML]{000000} 0.9254 & {\cellcolor[HTML]{B8C6E0}} \color[HTML]{000000} 0.9244 & {\cellcolor[HTML]{F7FCB4}} \color[HTML]{000000} 2.78\% & {\cellcolor[HTML]{FFFDBC}} \color[HTML]{000000} -0.58\% & {\cellcolor[HTML]{FFFEBE}} \color[HTML]{000000} -0.17\% & {\cellcolor[HTML]{F8FCB6}} \color[HTML]{000000} 2.36\% & {\cellcolor[HTML]{FAFDB8}} \color[HTML]{000000} 2.15\% \\
\hline
\end{tabular}
}
\caption{Comparison of detection performance between originally and adversarially trained MGT detection methods (in columns) on the original (unobfuscated) and per AO method (obfuscated) data (shown in rows). Mean across detection models absolute AUC ROC values are shown in the left part of the table. Relative difference in AUC ROC caused by adversarially trained methods in comparison to originally trained are shown in the right part of the table.}
\label{tab:exp-drop-at}
\end{table*}

The performance results of originally and adversarially trained MGT detection methods in \figurename~\ref{fig:fig-at} show that in general \textbf{adversarial retraining by data augmentation using obfuscated texts increases the overall performance} on the used test set. However, in some cases
of adversarially trained models, we observe a decrease in the overall performance.
To provide more insights, Table~\ref{tab:exp-drop-at} in Appendix~\ref{sec:appendix-results} shows the results per AO method (i.e., a portion of the test set containing obfuscated machine-class texts by a particular AO method). It provides absolute AUC ROC values as well as relative values reflecting the effect of adversarial retraining (i.e., whether it helped or not).

Table~\ref{tab:exp-drop-at} shows the results per AO method (i.e., a portion of the test set containing obfuscated machine-class texts by a particular AO method). It provides absolute AUC ROC values (mean across detection methods) to illustrate performance on different portions of the test set, as well as relative values reflecting the effect of adversarial retraining.
Using texts obfuscated by \textit{all} AO methods for adversarial retraining slightly decreases the performance on the original test data, but it significantly increases the performance for homoglyph attacks, having negligible impact on other portions of the test data.
Using \textit{m2m100-1.2B} and \textit{ChatGPT} increases performance for backtranslation and paraphrasing, but slightly drops the performance in other cases, especially \textit{HomoglyphAttack}.
Using \textit{HomoglyphAttack} significantly increases performance for homoglyph attacks, but slightly decreases performance in other cases.
And finally, using \textit{DFTFooler} increases performance for most of the text-edits attacks, but slightly drops performance in other cases.

\textbf{There are differences in the effect of adversarial retraining based on the used AO method data.} 
Using texts obfuscated by all AO methods is the best option on average, while using backtranslation only is the worst option (although comparable with ChatGPT basic paraphrasing). Adversarial retraining using single AO method data does not only increase the performance on that particular AO method obfuscated test data, but transfers also to some other obfuscated texts. For example, using adversarial attack also increased performance for homoglyph attacks. For additional per-language results, see Table~\ref{tab:exp-drop-at-diff} in Appendix~\ref{sec:appendix-results}.


\section{Discussion}
\label{sec:discussion}

\textbf{Existing authorship-obfuscation methods present a potential threat in non-English languages.} 
All of them are able to confuse all detection methods at least for some texts. There is a sole exception of GPTZeroBypasser not being able to fool RoBERTa-base-autextification-Detection in any test language (due to its rather random detection ability). 
Still, in majority of cases (combinations of AO method, MGT detection method, and test language), the attack success rate is below 50\%. However, we have not considered combinations of multiple AO methods or sophisticated adjustments to multilingualism. 

\textbf{Pre-processing to avoid homoglyphs can protect against this type of attacks.} 
Homoglyphs are usually considered as being able to fool humans due to visual similarity of characters but not the computers due to unrelated binary representations. If the classification models are not trained considering such cases, homoglyph-including samples can easily be misclassified. However, pre-processing steps specifically targeting homoglyphs (e.g., characters of different scripts contained in a single word) can presumably alleviate this kind of attacks. Data augmentation using such obfuscated samples also helps, as shown in our experiments.


\textbf{A simple data augmentation using obfuscated texts can increase adversarial robustness of MGT detection methods.} Usually, a complex adversarial training including generation of adversarial examples targeting a specific model is used to increase adversarial robustness of a model. We have shown that inclusion of generic obfuscated samples (not necessarily being adversarial examples) also helps to increase the adversarial robustness. We also observe a transferability between AO methods, meaning that using samples from one AO method for adversarial training increases resistance to another AO method during testing.

\section{Conclusion}
\label{sec:conclusion}

In this work, we have done the first comprehensive benchmark of authorship obfuscation methods in multilingual settings. We have also done the first evaluation of adversarial robustness (using authorship obfuscation methods) of multilingual machine-generated text detection methods. We have crafted a new public dataset of 740k obfuscated texts, which has wide usage possibilities, even beyond the purpose outlined in this work (e.g., to study differences in obfuscated human-written and obfuscated machine-generated texts, or to evaluate robustness of multi-class authorship-attribution systems).

Our results (namely nonzero attack success rate values) indicate that the existing authorship obfuscation methods, although focused primarily on English, are usable in all 11 tested languages, thus posing a potential threat for automated multilingual detection. However, human quality check of the obfuscated texts disqualified two of them as unusable in non-English and/or non-Latin script languages. 
The adversarial robustness results show that there are differences among machine-generated text detection method categories as well as among tested languages. The results confirmed that English monolingual fine-tuning of multilingual detectors is not suitable, since it is the least resistant against obfuscation.
As we have shown, the adversarial retraining using a simple data augmentation by obfuscated texts can significantly increase the adversarial robustness of detectors, especially against homoglyph and paraphrasing attacks.

Future work should be focused on iterative and hybrid authorship obfuscation of multilingual texts to evaluate severity of such threat on machine-generated text detection.

\section*{Limitations}
\label{sec:limitations}

\paragraph{Selection of AO Methods.} We have limited the scope of this work by constraining which \textbf{authorship obfuscation methods} (e.g., fully automated) have been used under which conditions, as described in Section~\ref{sec:obfuscation}. We have limited the number of trials of obfuscation methods to 10, leaving some texts not modified (i.e., they might have remained the same as the original). Due to computational and time costs, we have avoided paid services (e.g., Google Translation\footnote{\scriptsize\url{https://cloud.google.com/translate?hl=en}}) and slow adversarial attacks (e.g., TextFooler\footnote{\scriptsize\url{https://github.com/jind11/TextFooler}}). As already noted, in real settings, actors trying to evade the detection of machine-generated texts might use a combination of AO methods or use humans in the loop.

\paragraph{Settings of AO Methods.} We have used only the default settings of AO methods where available. By tuning the obfuscation settings, one could achieve slightly different results, as shown by our ablation study for HomoglyphAttack in Appendix~\ref{sec:appendix-ablation} (Table~\ref{tab:abl-asr-ao}). Therefore, a comparison between individual AO methods (e.g., the best and worst AO methods) should be interpreted with this in mind. However, general observations about multilingual usage of existing AO methods, per-language differences, or adversarial robustness evaluations are not significantly affected by these and still hold.

\paragraph{Dataset Selection.} We have used a \textbf{single dataset} (MULTITuDE) for experiments, limiting the number of \textbf{languages} for training (3) and for evaluation (11), limiting the number of \textbf{LLMs} used to generate the machine texts (8), and limiting the \textbf{domain} of the texts to news articles. Extension in all the mentioned aspects as well as in the \textbf{number of samples} (especially better balancing between human and machine classes) could help the generalizations; however, not influencing the conclusions we could already make.

\paragraph{Evaluation of Obfuscation Quality.} We evaluated the quality (usability) of the selected AO methods using a range of heuristics (i.e., similarity metrics, change in text length, language change, etc., see Appendix~\ref{sec:appendix-similarity}). Additionally, we did a manual quality check on a small random subset of the obfuscated data to evaluate their quality from the users' perspective, i.e., whether the obfuscation does not make the texts easily detectable by humans in the process. Nevertheless, we did not evaluate the quality on the whole dataset, which was unfeasible due to the scale of our dataset (740k obfuscated texts).

\section*{Ethics Statement}
\label{sec:ethics}

Our work identifies the most effective authorship obfuscation methods for each tested language, thus identifying potential vulnerabilities of machine-generated text detection methods. Although there is a risk that this can be misused by adversaries, the benefits of our work outweigh such a risk. First, we use existing freely available AO methods. Secondly, although we identify weaknesses of already existing detection methods, we also provide simple means to eliminate them through data augmentation during adversarial retraining.

We use an existing data resource in our work (MULTITuDE) in accordance with its intended use and license. As a part of our work, we extend this resource by generating obfuscated human and machine-generated texts. We will publish this data (for research purposes only) together with our code to ensure reproducibility of our work. 

\section*{Acknowledgements}
This work was partially supported by the projects funded by the European Union under the Horizon Europe: \textit{VIGILANT}, GA No. \href{https://doi.org/10.3030/101073921}{101073921}, \textit{AI-CODE}, GA No. \href{https://cordis.europa.eu/project/id/101135437}{101135437}; and by \textit{Modermed}, a project funded by the Slovak Research and Development Agency, GA No. APVV-22-0414.
This work was also in part supported by U.S. National Science Foundation (NSF) awards \#1820609, \#1934782, \#2114824, and \#2131144.

Part of the research results was obtained using the computational resources provided by the
CloudBank (https://www.cloudbank.org/), which
was supported by the NSF award \#1925001, and
those procured in the national project \textit{National competence centre for high performance computing} (project code: 311070AKF2) funded by European Regional Development Fund, EU Structural Funds Informatization of Society, Operational Program Integrated Infrastructure.

\bibliography{anthology,custom}

\appendix

\section{Computational Resources}
\label{sec:appendix-resources}

In the experiments, we have utilized a computational infrastructure consisting of 1$\times$ A100 40GB GPUs for application of the authorship obfuscation methods, text-similarity metrics calculations, and machine-generated text detectors requiring LLM inference and fine-tuning of pre-trained models. Cumulatively, these processes required approximately 1,875 GPU hours. For other processes, such as data analysis, simpler metric calculations (not requiring GPU), using some obfuscation methods (ChatGPT inference via OpenAI API, GPTZzzs, GPTZeroBypasser, and HomoglyphAttack), and results analysis, we used Google Colab\footnote{\scriptsize\url{https://colab.research.google.com/}} without GPU acceleration.

\section{Authorship Obfuscation Methods}
\label{sec:appendix-ao}

The used authorship obfuscation methods, grouped into three categories, are summarized in Table~\ref{tab:ao_methods} along with the used default parameters or settings for obfuscation.

\begin{savenotes}
\begin{table}[!h]
\footnotesize 
\centering
\resizebox{\linewidth}{!}{
\begin{tabular}{m{0.2cm}|p{3.2cm}@{}p{6cm}}
\hline
& \textbf{AO Method} & \textbf{Parameters} \\
\hline
\multirow[c]{6}{*}{\rotatebox{90}{\scriptsize\textbf{Backtranslation }}}
& m2m100-1.2B\footnote{\scriptsize\url{https://huggingface.co/facebook/m2m100_1.2B}}\newline \citep{fan2020englishcentric} & We have used English as an intermediary language for non-English texts and Spanish for English texts.\\
& nllb-200-distilled-1.3B\footnote{\scriptsize\url{https://huggingface.co/facebook/nllb-200-distilled-1.3B}}\newline \citep{nllbteam2022language} & We have used English as an intermediary language for non-English texts and Spanish for English texts, with $max\_length$ set to 512.\\
\hline
\multirow[c]{14}{*}{\rotatebox{90}{\textbf{Paraphrasing}}} & Pegasus-paraphrase\footnote{\scriptsize\url{https://huggingface.co/tuner007/pegasus_paraphrase}} & We have used the model for paraphrasing each sentence separately, with $max\_length$ of 60, $num\_beams$ of 10, and $temperature$ of 1.5, as provided in exemplar usage on HuggingFace.\\
& DIPPER\footnote{\scriptsize\url{https://github.com/martiansideofthemoon/ai-detection-paraphrases}}\newline \citep{krishna2023paraphrasing} & We have used both the $lex\_diversity$ and $order\_diversity$ set to 40 (as the most intensive settings in the DIPPER paper), the nucleus sampling with $top\_p$ of 0.75 and $max\_length$  of 512.\\
& ChatGPT\footnote{\scriptsize\url{https://openai.com/blog/chatgpt}} & We have used a basic paraphrasing prompt of \texttt{``Paraphrase the following text in <language> language: <text>''}. We have limited the number of output tokens to 512, and used the nucleus sampling with $top\_p$ of 0.95.\\
\hline
\multirow[c]{20}{*}{\rotatebox{90}{\textbf{Text edits}}} & GPTZzzs\footnote{\scriptsize\url{https://github.com/Declipsonator/GPTZzzs}} & We have used a random seed of 42. We have set 30\% of words to be replaced by synonyms using Zaibacu Thesaurus and 50\% of adjectives to changed emphasis on, without ignoring quotations and without using only common words. \\
& GPTZeroBypasser\footnote{\scriptsize\url{https://github.com/o2161405/GPTZero-Bypasser}} & We have used a random seed of 42, inserting a zero-width joiner pseudorandomly with a probability of 0.2.\\
& HomoglyphAttack & We have used a random seed of 42. We have used the whole confusables table\footnote{\scriptsize\url{https://www.unicode.org/Public/security/8.0.0/confusables.txt}} to psudorandomly replace letters for their homoglyphs with a probability (of a character being replaced) set to 0.1.\\
& ALISON\footnote{\scriptsize\url{https://github.com/EricX003/ALISON}}\newline \citep{alison} & We have use bert-base-multilingual-cased as a base model and $min\_length$ of 2, other parameters used the default values.\\
& DFTFooler\footnote{\scriptsize\url{https://github.com/jmpu/DeepfakeTextDetection/tree/main/DFTFooler}}\newline \citep{pu2023deepfake} & We have used bert-base-multilingual-cased as the backend model and the number of samples to attack set to 100,000 to obfuscate all texts. Other parameters used the default values.\\
\hline
\end{tabular}
}
\caption{Authorship obfuscation methods parameters.}
\label{tab:ao_methods}
\end{table}
\end{savenotes}

\begin{table*}[!t]
\centering
\resizebox{\linewidth}{!}{
\begin{tabular}{lcccccccc}
\hline
\bfseries AO Method & \bfseries METEOR $\uparrow$ & \bfseries BERTScore $\uparrow$ & \bfseries USE $\uparrow$ & \bfseries ngram $\uparrow$ & \bfseries TF $\uparrow$ & \bfseries LD $\downarrow$ & \bfseries CharLenDiff $\rightarrow 1$ & \bfseries LangCheck $\downarrow$ \\
\hline
\bfseries m2m100-1.2B & 0.452 (±0.22) & 0.853 (±0.07) & 0.842 (±0.13) & 0.485 (±0.18) & 0.810 (±0.16) & 0.467 (±0.21) & 0.678 (±0.24) & 0.55\% \\
\bfseries nllb-200-distilled-1.3B & 0.398 (±0.23) & 0.833 (±0.08) & 0.797 (±0.17) & 0.431 (±0.20) & 0.775 (±0.18) & 0.542 (±0.33) & 0.638 (±0.39) & \bfseries 0.30\% \\
\bfseries Pegasus-paraphrase & 0.331 (±0.24) & 0.708 (±0.15) & 0.575 (±0.34) & 0.324 (±0.23) & 0.646 (±0.28) & 0.698 (±0.40) & 0.556 (±0.49) & 28.17\% \\
\bfseries DIPPER & 0.276 (±0.23) & 0.760 (±0.10) & 0.683 (±0.26) & 0.282 (±0.23) & 0.528 (±0.34) & 0.704 (±0.28) & 0.756 (±0.32) & 51.79\% \\
\bfseries ChatGPT & 0.566 (±0.22) & 0.867 (±0.07) & 0.884 (±0.11) & 0.546 (±0.18) & 0.819 (±0.16) & 0.418 (±0.22) & 0.920 (±0.24) & 1.38\% \\
\bfseries GPTZzzs & 0.968 (±0.06) & 0.974 (±0.03) & 0.988 (±0.02) & 0.918 (±0.09) & \bfseries 0.986 (±0.02) & 0.046 (±0.05) & 1.017 (±0.02) & 2.78\% \\
\bfseries GPTZeroBypasser & 0.131 (±0.10) & 0.651 (±0.21) & 0.375 (±0.18) & 0.168 (±0.14) & 0.130 (±0.17) & 0.495 (±0.17) & 1.238 (±0.03) & 37.33\% \\
\bfseries HomoglyphAttack & 0.568 (±0.10) & 0.778 (±0.05) & 0.762 (±0.11) & 0.596 (±0.06) & 0.179 (±0.16) & 0.094 (±0.02) & \bfseries 1.003 (±0.00) & 2.74\% \\
\bfseries ALISON & \bfseries 0.987 (±0.06) & \bfseries 0.991 (±0.02) & \bfseries 0.993 (±0.01) & \bfseries 0.971 (±0.04) & 0.968 (±0.07) & \bfseries 0.009 (±0.01) & 1.005 (±0.01) & 2.77\% \\
\bfseries DFTFooler & 0.948 (±0.07) & 0.977 (±0.02) & 0.990 (±0.02) & 0.920 (±0.08) & 0.963 (±0.06) & 0.033 (±0.04) & 1.004 (±0.01) & 2.78\% \\
\hline
\end{tabular}
}
\caption{Similarity analysis between obfuscated and original texts [mean ($\pm$ std)]. \textit{LD} is normalized Levenshtein distance, \textit{LangCheck} is percentage of texts with changed languages based on FastText predictions, arrows refer to values representing more similar texts, boldfaced values represent the most similar texts for each metric.}
\label{tab:textanalysis}
\end{table*}

\begin{table*}[!t]
\centering
\resizebox{0.8\linewidth}{!}{
\begin{tabular}{lcccccccc}
\hline
\bfseries LLM Generator & \bfseries METEOR $\uparrow$ & \bfseries BERTScore $\uparrow$ & \bfseries USE $\uparrow$ & \bfseries ngram $\uparrow$ & \bfseries TF $\uparrow$ & \bfseries LangCheck $\downarrow$ \\
\hline
\bfseries alpaca-lora-30b & 0.110 (±0.07) & 0.668 (±0.04) & 0.516 (±0.20) & 0.170 (±0.08) & 0.619 (±0.20) & 1.01\%\\
\bfseries gpt-3.5-turbo & 0.139 (±0.07) & 0.678 (±0.04) & 0.584 (±0.20) & 0.215 (±0.09) & 0.650 (±0.20) & 0.02\%\\
\bfseries gpt-4 & \bfseries 0.163 (±0.08) & \bfseries 0.688 (±0.04) & \bfseries 0.629 (±0.21) & \bfseries 0.253 (±0.10) & \bfseries 0.667 (±0.21) & 0.00\%\\
\bfseries llama-65b & 0.099 (±0.07) & 0.619 (±0.06) & 0.448 (±0.22) & 0.138 (±0.10) & 0.513 (±0.23) & 14.29\%\\
\bfseries opt-66b & 0.116 (±0.08) & 0.655 (±0.05) & 0.464 (±0.25) & 0.175 (±0.10) & 0.595 (±0.23) & 3.53\%\\
\bfseries opt-iml-max-1.3b & 0.106 (±0.08) & 0.635 (±0.06) & 0.402 (±0.26) & 0.159 (±0.10) & 0.548 (±0.23) & 4.80\%\\
\bfseries text-davinci-003 & 0.123 (±0.07) & 0.674 (±0.04) & 0.542 (±0.21) & 0.196 (±0.09) & 0.620 (±0.21) & 0.00\%\\
\bfseries vicuna-13b & 0.131 (±0.07) & 0.667 (±0.04) & 0.548 (±0.21) & 0.199 (±0.09) & 0.630 (±0.21) & 2.89\%\\
\hline
\end{tabular}
}
\caption{Similarity analysis between machine-generated and human-written texts in original MULTITuDE data [mean ($\pm$ std)]. \textit{LangCheck} is percentage of texts with intended languages mismatching FastText predictions, arrows refer to values representing more similar texts, boldfaced values represent the most similar texts for each metric.}
\label{tab:generatedtextanalysis}
\end{table*}

\section{Post-Obfuscation Similarity Analysis}
\label{sec:appendix-similarity}

To evaluate the similarity between the obfuscated and the original texts (and thus the quality of the obfuscation), we have used the following metrics: \textit{METEOR} \citep{banerjee-lavie-2005-meteor} as a standard metric used in machine translation based on unigrams, \textit{BERTScore} \citep{zhang2019bertscore} with mBERT model as a contextual embeddings based similarity metric that is more robust to adversarial texts, and cosine similarity of \textit{USE} (multilingual version of universal sentence encoder) \citep{yang2019multilingual} sentence-level embeddings as a semantic similarity metric. For a better explainability, we have also used character-level \textit{ngram}\footnote{\scriptsize\url{https://pypi.org/project/ngram}} (3-grams in our case) as a language-independent string similarity metric in the form of a ratio of the shared ngrams between two strings and cosine similarity of \textit{TF} (term frequency) \citep{10.1145/775047.775110} as a word-level frequency similarity metric by using polyglot\footnote{\scriptsize\url{https://github.com/aboSamoor/polyglot}} multilingual tokenizer. All of the mentioned metrics represent a score between 0 and 1, while closer to 1 means higher similarity. In addition, we have measured amount (severity) of text modification by using the \textit{Levenshtein distance} as a character-level edit distance\footnote{\scriptsize\url{https://github.com/roy-ht/editdistance}}, which was normalized to the text length (a number closer to 0 means lesser modification). Since some LLMs occasionally fail in the task by generating duplicated characters, we have calculated a ratio of character-level lengths between obfuscated and unobfuscated texts after removal of subsequent duplicate characters (e.g., 'ssssssss' is changed to a single 's' for the calculation purpose). We call this ratio \textit{CharLenDiff}. Such a ratio value above 1 represents that the obfuscation lengthened the text and between 0 to 1 means that obfuscation shortened the text (while not taking duplicated characters into account). Also, since multiple of the used AO methods are English-only by their nature, we have checked possible changed language of the obfuscated texts by \textit{FastText}\footnote{\scriptsize\url{https://github.com/facebookresearch/fastText/}} language detection.

\begin{table*}[!t]
\centering
\resizebox{\linewidth}{!}{
\begin{tabular}{l|ccccccccccc}
\hline
 & \multicolumn{11}{c}{\bfseries Test Language [mean ($\pm$ std)]} \\
\bfseries AO Method & \bfseries ar & \bfseries ca & \bfseries cs & \bfseries de & \bfseries en & \bfseries es & \bfseries nl & \bfseries pt & \bfseries ru & \bfseries uk & \bfseries zh \\
\hline
\bfseries m2m100-1.2B & {\cellcolor[HTML]{C7E77F}} \color[HTML]{000000} 0.9 (±0.31) & {\cellcolor[HTML]{D9EF8B}} \color[HTML]{000000} 0.8 (±0.41) & {\cellcolor[HTML]{C3E67D}} \color[HTML]{000000} 0.9 (±0.25) & {\cellcolor[HTML]{D9EF8B}} \color[HTML]{000000} 0.8 (±0.41) & {\cellcolor[HTML]{E5F49B}} \color[HTML]{000000} 0.7 (±0.47) & {\cellcolor[HTML]{C7E77F}} \color[HTML]{000000} 0.9 (±0.31) & {\cellcolor[HTML]{DDF191}} \color[HTML]{000000} 0.8 (±0.43) & {\cellcolor[HTML]{C3E67D}} \color[HTML]{000000} 0.9 (±0.25) & {\cellcolor[HTML]{D3EC87}} \color[HTML]{000000} 0.8 (±0.53) & {\cellcolor[HTML]{B7E075}} \color[HTML]{000000} 1.0 (±0.00) & {\cellcolor[HTML]{E2F397}} \color[HTML]{000000} 0.7 (±0.52) \\
\bfseries nllb-200-distilled-1.3B & {\cellcolor[HTML]{B7E075}} \color[HTML]{000000} 1.0 (±0.00) & {\cellcolor[HTML]{EEF8A8}} \color[HTML]{000000} 0.6 (±0.72) & {\cellcolor[HTML]{F7FCB4}} \color[HTML]{000000} 0.6 (±0.68) & {\cellcolor[HTML]{EEF8A8}} \color[HTML]{000000} 0.6 (±0.76) & {\cellcolor[HTML]{DDF191}} \color[HTML]{000000} 0.8 (±0.43) & {\cellcolor[HTML]{FBFDBA}} \color[HTML]{000000} 0.5 (±0.73) & {\cellcolor[HTML]{FFFCBA}} \color[HTML]{000000} 0.5 (±0.78) & {\cellcolor[HTML]{DDF191}} \color[HTML]{000000} 0.8 (±0.57) & {\cellcolor[HTML]{DDF191}} \color[HTML]{000000} 0.8 (±0.50) & {\cellcolor[HTML]{C7E77F}} \color[HTML]{000000} 0.9 (±0.31) & {\cellcolor[HTML]{CDEA83}} \color[HTML]{000000} 0.9 (±0.35) \\
\bfseries Pegasus-paraphrase & {\cellcolor[HTML]{A50026}} \color[HTML]{F1F1F1} -1.0 (±0.00) & {\cellcolor[HTML]{FEE491}} \color[HTML]{000000} 0.2 (±0.68) & {\cellcolor[HTML]{FFFCBA}} \color[HTML]{000000} 0.5 (±0.51) & {\cellcolor[HTML]{FFF8B4}} \color[HTML]{000000} 0.4 (±0.63) & {\cellcolor[HTML]{D3EC87}} \color[HTML]{000000} 0.8 (±0.46) & {\cellcolor[HTML]{FEDA86}} \color[HTML]{000000} 0.2 (±0.95) & {\cellcolor[HTML]{FFF1A8}} \color[HTML]{000000} 0.4 (±0.61) & {\cellcolor[HTML]{FFF5AE}} \color[HTML]{000000} 0.4 (±0.77) & {\cellcolor[HTML]{A50026}} \color[HTML]{F1F1F1} -1.0 (±0.00) & {\cellcolor[HTML]{A50026}} \color[HTML]{F1F1F1} -1.0 (±0.00) & {\cellcolor[HTML]{A50026}} \color[HTML]{F1F1F1} -1.0 (±0.00) \\
\bfseries DIPPER & {\cellcolor[HTML]{BB1526}} \color[HTML]{F1F1F1} -0.9 (±0.51) & {\cellcolor[HTML]{D62F27}} \color[HTML]{F1F1F1} -0.7 (±0.65) & {\cellcolor[HTML]{BB1526}} \color[HTML]{F1F1F1} -0.9 (±0.43) & {\cellcolor[HTML]{A90426}} \color[HTML]{F1F1F1} -1.0 (±0.18) & {\cellcolor[HTML]{FBFDBA}} \color[HTML]{000000} 0.5 (±0.78) & {\cellcolor[HTML]{ED5F3C}} \color[HTML]{F1F1F1} -0.5 (±0.82) & {\cellcolor[HTML]{BB1526}} \color[HTML]{F1F1F1} -0.9 (±0.35) & {\cellcolor[HTML]{BB1526}} \color[HTML]{F1F1F1} -0.9 (±0.51) & {\cellcolor[HTML]{A50026}} \color[HTML]{F1F1F1} -1.0 (±0.00) & {\cellcolor[HTML]{A50026}} \color[HTML]{F1F1F1} -1.0 (±0.00) & {\cellcolor[HTML]{A50026}} \color[HTML]{F1F1F1} -1.0 (±0.00) \\
\bfseries ChatGPT & {\cellcolor[HTML]{B7E075}} \color[HTML]{000000} 1.0 (±0.00) & {\cellcolor[HTML]{C7E77F}} \color[HTML]{000000} 0.9 (±0.40) & {\cellcolor[HTML]{CDEA83}} \color[HTML]{000000} 0.9 (±0.51) & {\cellcolor[HTML]{F7FCB4}} \color[HTML]{000000} 0.6 (±0.82) & {\cellcolor[HTML]{C3E67D}} \color[HTML]{000000} 0.9 (±0.37) & {\cellcolor[HTML]{B7E075}} \color[HTML]{000000} 1.0 (±0.00) & {\cellcolor[HTML]{B7E075}} \color[HTML]{000000} 1.0 (±0.00) & {\cellcolor[HTML]{B7E075}} \color[HTML]{000000} 1.0 (±0.00) & {\cellcolor[HTML]{C3E67D}} \color[HTML]{000000} 0.9 (±0.37) & {\cellcolor[HTML]{B7E075}} \color[HTML]{000000} 1.0 (±0.00) & {\cellcolor[HTML]{DDF191}} \color[HTML]{000000} 0.8 (±0.63) \\
\bfseries GPTZzzs & {\cellcolor[HTML]{B7E075}} \color[HTML]{000000} 1.0 (±0.00) & {\cellcolor[HTML]{B7E075}} \color[HTML]{000000} 1.0 (±0.00) & {\cellcolor[HTML]{B7E075}} \color[HTML]{000000} 1.0 (±0.00) & {\cellcolor[HTML]{B7E075}} \color[HTML]{000000} 1.0 (±0.00) & {\cellcolor[HTML]{B7E075}} \color[HTML]{000000} 1.0 (±0.00) & {\cellcolor[HTML]{B7E075}} \color[HTML]{000000} 1.0 (±0.00) & {\cellcolor[HTML]{B7E075}} \color[HTML]{000000} 1.0 (±0.00) & {\cellcolor[HTML]{B7E075}} \color[HTML]{000000} 1.0 (±0.00) & {\cellcolor[HTML]{B7E075}} \color[HTML]{000000} 1.0 (±0.00) & {\cellcolor[HTML]{B7E075}} \color[HTML]{000000} 1.0 (±0.00) & {\cellcolor[HTML]{B7E075}} \color[HTML]{000000} 1.0 (±0.00) \\
\bfseries GPTZeroBypasser & {\cellcolor[HTML]{B7E075}} \color[HTML]{000000} 1.0 (±0.00) & {\cellcolor[HTML]{E9F6A1}} \color[HTML]{000000} 0.7 (±0.48) & {\cellcolor[HTML]{E9F6A1}} \color[HTML]{000000} 0.7 (±0.48) & {\cellcolor[HTML]{E9F6A1}} \color[HTML]{000000} 0.7 (±0.48) & {\cellcolor[HTML]{E9F6A1}} \color[HTML]{000000} 0.7 (±0.48) & {\cellcolor[HTML]{E9F6A1}} \color[HTML]{000000} 0.7 (±0.48) & {\cellcolor[HTML]{F7FCB4}} \color[HTML]{000000} 0.6 (±0.50) & {\cellcolor[HTML]{EEF8A8}} \color[HTML]{000000} 0.6 (±0.49) & {\cellcolor[HTML]{C3E67D}} \color[HTML]{000000} 0.9 (±0.25) & {\cellcolor[HTML]{CDEA83}} \color[HTML]{000000} 0.9 (±0.35) & {\cellcolor[HTML]{EEF8A8}} \color[HTML]{000000} 0.6 (±0.49) \\
\bfseries HomoglyphAttack & {\cellcolor[HTML]{FEEDA1}} \color[HTML]{000000} 0.3 (±0.48) & {\cellcolor[HTML]{FEEDA1}} \color[HTML]{000000} 0.3 (±0.48) & {\cellcolor[HTML]{FEEDA1}} \color[HTML]{000000} 0.3 (±0.48) & {\cellcolor[HTML]{FEEDA1}} \color[HTML]{000000} 0.3 (±0.48) & {\cellcolor[HTML]{FEEDA1}} \color[HTML]{000000} 0.3 (±0.48) & {\cellcolor[HTML]{FEEDA1}} \color[HTML]{000000} 0.3 (±0.48) & {\cellcolor[HTML]{FEEDA1}} \color[HTML]{000000} 0.3 (±0.48) & {\cellcolor[HTML]{FEEDA1}} \color[HTML]{000000} 0.3 (±0.48) & {\cellcolor[HTML]{FEEDA1}} \color[HTML]{000000} 0.3 (±0.48) & {\cellcolor[HTML]{FFF1A8}} \color[HTML]{000000} 0.4 (±0.49) & {\cellcolor[HTML]{F2FAAE}} \color[HTML]{000000} 0.6 (±0.50) \\
\bfseries ALISON & {\cellcolor[HTML]{B7E075}} \color[HTML]{000000} 1.0 (±0.00) & {\cellcolor[HTML]{B7E075}} \color[HTML]{000000} 1.0 (±0.00) & {\cellcolor[HTML]{B7E075}} \color[HTML]{000000} 1.0 (±0.00) & {\cellcolor[HTML]{B7E075}} \color[HTML]{000000} 1.0 (±0.00) & {\cellcolor[HTML]{B7E075}} \color[HTML]{000000} 1.0 (±0.00) & {\cellcolor[HTML]{B7E075}} \color[HTML]{000000} 1.0 (±0.00) & {\cellcolor[HTML]{B7E075}} \color[HTML]{000000} 1.0 (±0.00) & {\cellcolor[HTML]{B7E075}} \color[HTML]{000000} 1.0 (±0.00) & {\cellcolor[HTML]{B7E075}} \color[HTML]{000000} 1.0 (±0.00) & {\cellcolor[HTML]{B7E075}} \color[HTML]{000000} 1.0 (±0.00) & {\cellcolor[HTML]{B7E075}} \color[HTML]{000000} 1.0 (±0.00) \\
\bfseries DFTFooler & {\cellcolor[HTML]{B7E075}} \color[HTML]{000000} 1.0 (±0.00) & {\cellcolor[HTML]{B7E075}} \color[HTML]{000000} 1.0 (±0.00) & {\cellcolor[HTML]{B7E075}} \color[HTML]{000000} 1.0 (±0.00) & {\cellcolor[HTML]{B7E075}} \color[HTML]{000000} 1.0 (±0.00) & {\cellcolor[HTML]{B7E075}} \color[HTML]{000000} 1.0 (±0.00) & {\cellcolor[HTML]{B7E075}} \color[HTML]{000000} 1.0 (±0.00) & {\cellcolor[HTML]{B7E075}} \color[HTML]{000000} 1.0 (±0.00) & {\cellcolor[HTML]{B7E075}} \color[HTML]{000000} 1.0 (±0.00) & {\cellcolor[HTML]{B7E075}} \color[HTML]{000000} 1.0 (±0.00) & {\cellcolor[HTML]{B7E075}} \color[HTML]{000000} 1.0 (±0.00) & {\cellcolor[HTML]{B7E075}} \color[HTML]{000000} 1.0 (±0.00) \\
\hline
\end{tabular}
}
\caption{Human quality check of the obfuscated data subset. Mean value for each AO method and each test language is reported, where -1.0 is the lowest quality (a red-like background color gradient) and 1.0 is the highest quality (a green-like background color gradient).}
\label{tab:humancheck}
\end{table*}

A summary of the post-obfuscation analysis is provided in Table~\ref{tab:textanalysis}. For a reference, a summary of per-generator similarity analysis between the corresponding human-written and machine-generated texts in the original MULTITuDE dataset is provided in Table~\ref{tab:generatedtextanalysis}.
Based on the results, we can see that the texts obfuscated by adversarial attacks are the most similar to the original versions. On the other hand, English-only paraphrasers (Pegasus-paraphrase and DIPPER) are clearly the outliers (by most of the metrics). As \textit{LangCheck} indicates, these methods probably changed the language of the texts to English (or incorporated several English parts), which must be considered during the experiments. Based on \textit{CharLenDiff}, we might assume that Pegasus-paraphrase AO method shortened the obfuscated texts by a high amount on average due to failing of the model to generate a meaningful text (shorter strings of duplicated characters mentioned above). Also, there is indicated a severe language change for GPTZeroBypasser AO method. In this case, however, we assume that FastText language detection was confused due to a change of portion of Latin characters to Cyrillic, while in HomoglyphAttack, the distribution of homoglyphs using different scripts is not as uniform (thus not causing the same effect). Homoglyph attacks achieve lower values for most of the similarity metrics, which is expected since tokens as well as ngrams are affected by changed characters (in contrast to humans being possibly fooled visually by such texts). Backtranslation and paraphrasing AO methods behave similarly in this analysis, both categories tending to shorten the texts (ChatGPT being exception, where shortening is not as severe).

\begin{table*}[!t]
\centering
\resizebox{\linewidth}{!}{
\begin{tabular}{p{3.7cm}|p{16cm}|c|p{3cm}}
\hline
\bfseries AO Method & \bfseries Obfuscated Text & \bfseries Quality & \bfseries Comment\\
\hline
\bfseries original & {\cellcolor{white}} Los dos soldados franceses que permanecían ingresados en el Hospital Universitario de Albacete tras ... & {\cellcolor{white}} N/A \\
\bfseries m2m100-1.2B & {\cellcolor[HTML]{B7E075}} Los muertos dos soldados franceses que quedaron en el Hospital Universitario de Albacete tras el acc... & {\cellcolor[HTML]{B7E075}} 1 \\
\bfseries nllb-200-distilled-1.3B & {\cellcolor[HTML]{B7E075}} Los dos soldados franceses que fueron ingresados en el Hospital Universitario de Albacete después de... & {\cellcolor[HTML]{B7E075}} 1 \\
\bfseries Pegasus-paraphrase & {\cellcolor[HTML]{B7E075}} The hospital Universitario de Albacete tras el accidente el pasado lunes de un F-16, ha sido traslad... & {\cellcolor[HTML]{B7E075}} 1 \\
\bfseries DIPPER & {\cellcolor[HTML]{DB382B}}  Most importantly, the Italian minister of defense visited the wounded in the hospital in Los Llanos... & {\cellcolor[HTML]{DB382B}} -1 & language change\\
\bfseries ChatGPT & {\cellcolor[HTML]{B7E075}} Los dos soldados franceses que estaban hospitalizados en el Hospital Universitario de Albacete despu... & {\cellcolor[HTML]{B7E075}} 1 \\
\bfseries GPTZzzs & {\cellcolor[HTML]{B7E075}} Los dos soldados franceses que permanecían ingresados en el Hospital Universitario de Albacete tras ... & {\cellcolor[HTML]{B7E075}} 1 \\
\bfseries GPTZeroBypasser & {\cellcolor[HTML]{FDBF6F}} L\ensuremath{\mathtt{o}}s d\ensuremath{\mathtt{o}}s s\ensuremath{\mathtt{o}}\ensuremath{\mathtt{l}}\,d\ensuremath{\mathtt{a}}d\ensuremath{\mathtt{o}}s fr\ensuremath{\mathtt{a}}\ensuremath{\mathtt{n}}\ensuremath{\mathtt{c}}eses \ensuremath{\mathtt{q}}\ensuremath{\mathtt{u}}e per\ensuremath{\mathtt{m}}\ensuremath{\mathtt{a}}\ensuremath{\mathtt{n}}e\ensuremath{\mathtt{c}}{\'\i}\ensuremath{\mathtt{a}}\ensuremath{\mathtt{n}} i\ensuremath{\mathtt{n}}gres\ensuremath{\mathtt{a}}d\ensuremath{\mathtt{o}}s e\ensuremath{\mathtt{n}} e\ensuremath{\mathtt{l}}\, H\ensuremath{\mathtt{o}}spi\ensuremath{\mathtt{t}}\ensuremath{\mathtt{a}}\ensuremath{\mathtt{l}}\, U\ensuremath{\mathtt{n}}iversi\ensuremath{\mathtt{t}}\ensuremath{\mathtt{a}}ri\ensuremath{\mathtt{o}}... & {\cellcolor[HTML]{FDBF6F}} 0 & multiple scripts\\
\bfseries HomoglyphAttack & {\cellcolor[HTML]{FDBF6F}} Los dos soldados franceses \ensuremath{\mathsf{Q}}ue permaec{\'\i}n ingesados en el H\"{\~o}spi\v{t}al Uni\ensuremath{\bigvee}er\ensuremath{\boldsymbol{s}}itario de Abac\c{e}\ensuremath{\boldsymbol{\mathit{t}}}e tr\={\"a}s ... & {\cellcolor[HTML]{FDBF6F}} 0 & weird characters\\
\bfseries ALISON & {\cellcolor[HTML]{B7E075}} Los dos soldados franceses que permanecían ingresados en el Hospital Universitario de Albacete tras ... & {\cellcolor[HTML]{B7E075}} 1 \\
\bfseries DFTFooler & {\cellcolor[HTML]{B7E075}} Los dos soldados franceses que permanecían ingresados en el Hospital Universitario de Albacete tras ... & {\cellcolor[HTML]{B7E075}} 1 \\
\hline
\end{tabular}
}
\caption{An example of original and obfuscated sample by each AO method along with a human annotation.}
\label{tab:obfuscationexamples}
\end{table*}

\subsection{Human Quality Check}
The aim of the human quality check is to evaluate the obfuscated data quality, i.e., whether the obfuscated data is/is not clearly damaged from a human view. Massive human study is infeasible due to the scope and complexity of the work (it would require proficient annotators in 11 languages); therefore, a manual quality check by three human evaluators (selected from the paper authors) has been executed to identify unusable AO methods.

To reduce load and noise by LLM generators, we use just the obfuscated human data for AO evaluation (except for DFTFooler, which obfuscates only machine-class samples, where GPT-4 generated data are used instead). We have pseudo-randomly selected 10 samples per each test language for each AO method, which resulted in N = 1100 samples.
The samples were annotated on the 3 step scale:
\begin{itemize}
    \item \textit{-1 -- unusable text} -- no meaningful text at all (e.g., just some numbers, or repeated/random characters), text completely in a different language than intended, massive repetition (more than one duplicated word)
    \item \textit{0 -- obviously damaged but readable text} -- mixed languages (partly in the intended language), missing letters (e.g., letters with accents), mixed script (changed characters)
    \item \textit{1 -- no obvious damage in the text} -- no obvious modification unusual for a human text
\end{itemize}

The annotations by the three annotators achieved averaged pairwise Pearson correlation of 0.7. The total annotation agreement accuracy of 0.65 (i.e., full agreement -- 65\% of the samples were annotated by the same values by all three reviewers). When considering majority agreement (i.e., 2 of 3 annotations matched for a given sample), accuracy achieved 0.99.

The mean quality annotation values are provided in Table~\ref{tab:humancheck} per each AO method and for each test language. Pegasus-paraphrase and DIPPER AO methods are clearly disqualified form being usable for some non-English languages. Also, homoglyph attacks are identifiable by humans; however, if not used intensively, they can deceive a human reader.

An example of obfuscated texts by each AO method with a corresponding annotation of the quality and explanation in a comment is provided in Table~\ref{tab:obfuscationexamples}.

\section{Ablation Study}
\label{sec:appendix-ablation}

\begin{table*}[!t]
\centering
\resizebox{\linewidth}{!}{
\begin{tabular}{p{4cm}|p{1.2cm}p{1.2cm}p{1.2cm}p{1.2cm}p{1.2cm}p{1.2cm}p{1.2cm}p{1.2cm}p{1.2cm}p{1.2cm}p{1.2cm}|p{2cm}}
\hline
 & \multicolumn{11}{c|}{\bfseries Test Language [mean (±confidence interval)]} \\
\bfseries AO Method & \bfseries ar & \bfseries ca & \bfseries cs & \bfseries de & \bfseries en & \bfseries es & \bfseries nl & \bfseries pt & \bfseries ru & \bfseries uk & \bfseries zh & \bfseries $\rightarrow$ Average \\
\hline
\bfseries m2m100-1.2B & {\cellcolor[HTML]{BBE278}} \textcolor{black}{0.3177 (±0.12)} & {\cellcolor[HTML]{BDE379}} \textcolor{black}{0.3101 (±0.11)} & {\cellcolor[HTML]{D9EF8B}} \textcolor{black}{0.2008 (±0.12)} & {\cellcolor[HTML]{C7E77F}} \textcolor{black}{0.2671 (±0.10)} & {\cellcolor[HTML]{C7E77F}} \textcolor{black}{0.2727 (±0.11)} & {\cellcolor[HTML]{CFEB85}} \textcolor{black}{0.2389 (±0.11)} & {\cellcolor[HTML]{C5E67E}} \textcolor{black}{0.2764 (±0.10)} & {\cellcolor[HTML]{CDEA83}} \textcolor{black}{0.2495 (±0.11)} & {\cellcolor[HTML]{C7E77F}} \textcolor{black}{0.2729 (±0.11)} & {\cellcolor[HTML]{CBE982}} \textcolor{black}{0.2510 (±0.11)} & {\cellcolor[HTML]{C1E57B}} \textcolor{black}{0.2911 (±0.09)} & {\cellcolor[HTML]{C7E77F}} \textcolor{black}{0.2680} \\
\bfseries nllb-200-distilled-1.3B & {\cellcolor[HTML]{C3E67D}} \textcolor{black}{0.2890 (±0.04)} & {\cellcolor[HTML]{B3DF72}} \textcolor{black}{0.3475 (±0.03)} & {\cellcolor[HTML]{CDEA83}} \textcolor{black}{0.2492 (±0.04)} & {\cellcolor[HTML]{CFEB85}} \textcolor{black}{0.2369 (±0.04)} & {\cellcolor[HTML]{D9EF8B}} \textcolor{black}{0.1985 (±0.05)} & {\cellcolor[HTML]{D3EC87}} \textcolor{black}{0.2195 (±0.04)} & {\cellcolor[HTML]{CBE982}} \textcolor{black}{0.2564 (±0.03)} & {\cellcolor[HTML]{D1EC86}} \textcolor{black}{0.2297 (±0.04)} & {\cellcolor[HTML]{C7E77F}} \textcolor{black}{0.2721 (±0.04)} & {\cellcolor[HTML]{BFE47A}} \textcolor{black}{0.3025 (±0.06)} & {\cellcolor[HTML]{B7E075}} \textcolor{black}{0.3350 (±0.02)} & {\cellcolor[HTML]{C7E77F}} \textcolor{black}{0.2669} \\
\bfseries Pegasus-paraphrase & {\cellcolor[HTML]{B9E176}} \textcolor{black}{0.3259 (±0.09)} & {\cellcolor[HTML]{8CCD67}} \textcolor{black}{0.4803 (±0.05)} & {\cellcolor[HTML]{A0D669}} \textcolor{black}{0.4151 (±0.05)} & {\cellcolor[HTML]{9BD469}} \textcolor{black}{0.4367 (±0.06)} & {\cellcolor[HTML]{DFF293}} \textcolor{black}{0.1708 (±0.06)} & {\cellcolor[HTML]{96D268}} \textcolor{black}{0.4469 (±0.05)} & {\cellcolor[HTML]{93D168}} \textcolor{black}{0.4542 (±0.05)} & {\cellcolor[HTML]{7AC665}} \textcolor{black}{0.5319 (±0.04)} & {\cellcolor[HTML]{C1E57B}} \textcolor{black}{0.2898 (±0.08)} & {\cellcolor[HTML]{D3EC87}} \textcolor{black}{0.2228 (±0.06)} & {\cellcolor[HTML]{CFEB85}} \textcolor{black}{0.2371 (±0.08)} & {\cellcolor[HTML]{AFDD70}} \textcolor{black}{0.3647} \\
\bfseries DIPPER & {\cellcolor[HTML]{AFDD70}} \textcolor{black}{0.3614 (±0.08)} & {\cellcolor[HTML]{DCF08F}} \textcolor{black}{0.1842 (±0.05)} & {\cellcolor[HTML]{D7EE8A}} \textcolor{black}{0.2096 (±0.06)} & {\cellcolor[HTML]{D5ED88}} \textcolor{black}{0.2151 (±0.05)} & {\cellcolor[HTML]{E5F49B}} \textcolor{black}{0.1382 (±0.04)} & {\cellcolor[HTML]{D5ED88}} \textcolor{black}{0.2148 (±0.05)} & {\cellcolor[HTML]{DCF08F}} \textcolor{black}{0.1853 (±0.04)} & {\cellcolor[HTML]{D3EC87}} \textcolor{black}{0.2262 (±0.04)} & {\cellcolor[HTML]{D1EC86}} \textcolor{black}{0.2314 (±0.07)} & {\cellcolor[HTML]{CFEB85}} \textcolor{black}{0.2418 (±0.07)} & {\cellcolor[HTML]{B3DF72}} \textcolor{black}{0.3444 (±0.09)} & {\cellcolor[HTML]{D1EC86}} \textcolor{black}{0.2320} \\
\bfseries ChatGPT & {\cellcolor[HTML]{CFEB85}} \textcolor{black}{0.2355 (±0.04)} & {\cellcolor[HTML]{DAF08D}} \textcolor{black}{0.1878 (±0.07)} & {\cellcolor[HTML]{DCF08F}} \textcolor{black}{0.1865 (±0.06)} & {\cellcolor[HTML]{DCF08F}} \textcolor{black}{0.1801 (±0.09)} & {\cellcolor[HTML]{E8F59F}} \textcolor{black}{0.1211 (±0.10)} & {\cellcolor[HTML]{E3F399}} \textcolor{black}{0.1426 (±0.09)} & {\cellcolor[HTML]{DDF191}} \textcolor{black}{0.1735 (±0.09)} & {\cellcolor[HTML]{E2F397}} \textcolor{black}{0.1528 (±0.09)} & {\cellcolor[HTML]{D5ED88}} \textcolor{black}{0.2154 (±0.07)} & {\cellcolor[HTML]{D7EE8A}} \textcolor{black}{0.2044 (±0.07)} & {\cellcolor[HTML]{D7EE8A}} \textcolor{black}{0.2095 (±0.02)} & {\cellcolor[HTML]{DCF08F}} \textcolor{black}{0.1826} \\
\bfseries GPTZzzs & {\cellcolor[HTML]{F5FBB2}} \textcolor{black}{0.0481 (±0.10)} & {\cellcolor[HTML]{EFF8AA}} \textcolor{black}{0.0837 (±0.05)} & {\cellcolor[HTML]{F8FCB6}} \textcolor{black}{0.0362 (±0.08)} & {\cellcolor[HTML]{F2FAAE}} \textcolor{black}{0.0629 (±0.09)} & {\cellcolor[HTML]{B9E176}} \textcolor{black}{0.3269 (±0.04)} & {\cellcolor[HTML]{E9F6A1}} \textcolor{black}{0.1094 (±0.09)} & {\cellcolor[HTML]{EEF8A8}} \textcolor{black}{0.0860 (±0.06)} & {\cellcolor[HTML]{E8F59F}} \textcolor{black}{0.1181 (±0.09)} & {\cellcolor[HTML]{FAFDB8}} \textcolor{black}{0.0296 (±0.08)} & {\cellcolor[HTML]{FAFDB8}} \textcolor{black}{0.0300 (±0.11)} & {\cellcolor[HTML]{F8FCB6}} \textcolor{black}{0.0387 (±0.09)} & {\cellcolor[HTML]{EEF8A8}} \textcolor{black}{0.0882} \\
\bfseries GPTZeroBypasser & {\cellcolor[HTML]{78C565}} \textcolor{black}{0.5392 (±0.13)} & {\cellcolor[HTML]{87CB67}} \textcolor{black}{0.4947 (±0.14)} & {\cellcolor[HTML]{ADDC6F}} \textcolor{black}{0.3705 (±0.13)} & {\cellcolor[HTML]{78C565}} \textcolor{black}{0.5409 (±0.11)} & {\cellcolor[HTML]{60BA62}} \textcolor{black}{0.6147 (±0.14)} & {\cellcolor[HTML]{7AC665}} \textcolor{black}{0.5374 (±0.13)} & {\cellcolor[HTML]{70C164}} \textcolor{black}{0.5696 (±0.13)} & {\cellcolor[HTML]{66BD63}} \textcolor{black}{0.5947 (±0.13)} & {\cellcolor[HTML]{7FC866}} \textcolor{black}{0.5170 (±0.13)} & {\cellcolor[HTML]{8CCD67}} \textcolor{black}{0.4818 (±0.13)} & {\cellcolor[HTML]{BDE379}} \textcolor{black}{0.3095 (±0.10)} & {\cellcolor[HTML]{84CA66}} \textcolor{black}{0.5064} \\
\bfseries HomoglyphAttack & {\cellcolor[HTML]{6EC064}} \textcolor{black}{0.5723 (±0.04)} & {\cellcolor[HTML]{51B35E}} \textcolor{black}{0.6496 (±0.04)} & {\cellcolor[HTML]{89CC67}} \textcolor{black}{0.4868 (±0.02)} & {\cellcolor[HTML]{5AB760}} \textcolor{black}{0.6305 (±0.04)} & {\cellcolor[HTML]{45AD5B}} \textcolor{black}{0.6839 (±0.11)} & {\cellcolor[HTML]{5AB760}} \textcolor{black}{0.6324 (±0.05)} & {\cellcolor[HTML]{3FAA59}} \textcolor{black}{0.7009 (±0.03)} & {\cellcolor[HTML]{4BB05C}} \textcolor{black}{0.6687 (±0.05)} & {\cellcolor[HTML]{75C465}} \textcolor{black}{0.5520 (±0.03)} & {\cellcolor[HTML]{78C565}} \textcolor{black}{0.5449 (±0.03)} & {\cellcolor[HTML]{CBE982}} \textcolor{black}{0.2532 (±0.02)} & {\cellcolor[HTML]{6BBF64}} \textcolor{black}{0.5796} \\
\bfseries ALISON & {\cellcolor[HTML]{F1F9AC}} \textcolor{black}{0.0740 (±0.07)} & {\cellcolor[HTML]{F2FAAE}} \textcolor{black}{0.0679 (±0.06)} & {\cellcolor[HTML]{F2FAAE}} \textcolor{black}{0.0625 (±0.06)} & {\cellcolor[HTML]{F2FAAE}} \textcolor{black}{0.0700 (±0.05)} & {\cellcolor[HTML]{ECF7A6}} \textcolor{black}{0.0945 (±0.04)} & {\cellcolor[HTML]{F1F9AC}} \textcolor{black}{0.0705 (±0.05)} & {\cellcolor[HTML]{F4FAB0}} \textcolor{black}{0.0582 (±0.06)} & {\cellcolor[HTML]{F1F9AC}} \textcolor{black}{0.0751 (±0.05)} & {\cellcolor[HTML]{EEF8A8}} \textcolor{black}{0.0867 (±0.08)} & {\cellcolor[HTML]{E8F59F}} \textcolor{black}{0.1208 (±0.07)} & {\cellcolor[HTML]{F8FCB6}} \textcolor{black}{0.0337 (±0.07)} & {\cellcolor[HTML]{F1F9AC}} \textcolor{black}{0.0740} \\
\bfseries DFTFooler & {\cellcolor[HTML]{F7FCB4}} \textcolor{black}{0.0459 (±0.10)} & {\cellcolor[HTML]{D7EE8A}} \textcolor{black}{0.2070 (±0.11)} & {\cellcolor[HTML]{E5F49B}} \textcolor{black}{0.1376 (±0.12)} & {\cellcolor[HTML]{B9E176}} \textcolor{black}{0.3230 (±0.12)} & {\cellcolor[HTML]{D1EC86}} \textcolor{black}{0.2289 (±0.05)} & {\cellcolor[HTML]{D1EC86}} \textcolor{black}{0.2299 (±0.12)} & {\cellcolor[HTML]{C3E67D}} \textcolor{black}{0.2841 (±0.12)} & {\cellcolor[HTML]{C5E67E}} \textcolor{black}{0.2810 (±0.11)} & {\cellcolor[HTML]{E0F295}} \textcolor{black}{0.1616 (±0.11)} & {\cellcolor[HTML]{E3F399}} \textcolor{black}{0.1474 (±0.10)} & {\cellcolor[HTML]{F7FCB4}} \textcolor{black}{0.0397 (±0.08)} & {\cellcolor[HTML]{DAF08D}} \textcolor{black}{0.1896} \\

\hline
\bfseries $\downarrow$ Average & {\cellcolor[HTML]{C5E67E}} \textcolor{black}{0.2809} & {\cellcolor[HTML]{BFE47A}} \textcolor{black}{0.3013} & {\cellcolor[HTML]{CFEB85}} \textcolor{black}{0.2355} & {\cellcolor[HTML]{C1E57B}} \textcolor{black}{0.2963} & {\cellcolor[HTML]{C3E67D}} \textcolor{black}{0.2850} & {\cellcolor[HTML]{C3E67D}} \textcolor{black}{0.2842} & {\cellcolor[HTML]{BFE47A}} \textcolor{black}{0.3045} & {\cellcolor[HTML]{BBE278}} \textcolor{black}{0.3128} & {\cellcolor[HTML]{C9E881}} \textcolor{black}{0.2628} & {\cellcolor[HTML]{CBE982}} \textcolor{black}{0.2547} & {\cellcolor[HTML]{D7EE8A}} \textcolor{black}{0.2092} &  \\
\hline
\end{tabular}
}
\caption{Ablation of attack success rate of AO methods based on classification thresholds of individual MGT detection methods optimized 
 for each language separately. Per-language mean values of all detection methods are reported along with 95\% confidence interval error bounds.
 }
\label{tab:abl-asr-all}
\end{table*}

\begin{table*}[!t]
\centering
\resizebox{\linewidth}{!}{
\begin{tabular}{l|ccccccccccc|c}
\hline
 & \multicolumn{11}{c|}{\bfseries Test Language [mean]} \\
\bfseries AO Method & \bfseries ar & \bfseries ca & \bfseries cs & \bfseries de & \bfseries en & \bfseries es & \bfseries nl & \bfseries pt & \bfseries ru & \bfseries uk & \bfseries zh & \bfseries $\rightarrow$ Average \\
\hline
\bfseries m2m100-1.2B & {\cellcolor[HTML]{ECF7A6}} \color[HTML]{000000} 0.0980 & {\cellcolor[HTML]{CDEA83}} \color[HTML]{000000} 0.2455 & {\cellcolor[HTML]{E6F59D}} \color[HTML]{000000} 0.1281 & {\cellcolor[HTML]{D1EC86}} \color[HTML]{000000} 0.2295 & {\cellcolor[HTML]{CBE982}} \color[HTML]{000000} 0.2545 & {\cellcolor[HTML]{DDF191}} \color[HTML]{000000} 0.1738 & {\cellcolor[HTML]{CDEA83}} \color[HTML]{000000} 0.2474 & {\cellcolor[HTML]{D7EE8A}} \color[HTML]{000000} 0.2062 & {\cellcolor[HTML]{DFF293}} \color[HTML]{000000} 0.1718 & {\cellcolor[HTML]{E3F399}} \color[HTML]{000000} 0.1419 & {\cellcolor[HTML]{E9F6A1}} \color[HTML]{000000} 0.1155 & {\cellcolor[HTML]{DCF08F}} \color[HTML]{000000} 0.1829 \\
\bfseries nllb-200-distilled-1.3B & {\cellcolor[HTML]{ECF7A6}} \color[HTML]{000000} 0.0954 & {\cellcolor[HTML]{C5E67E}} \color[HTML]{000000} 0.2792 & {\cellcolor[HTML]{E0F295}} \color[HTML]{000000} 0.1589 & {\cellcolor[HTML]{DDF191}} \color[HTML]{000000} 0.1756 & {\cellcolor[HTML]{DFF293}} \color[HTML]{000000} 0.1670 & {\cellcolor[HTML]{E5F49B}} \color[HTML]{000000} 0.1338 & {\cellcolor[HTML]{D7EE8A}} \color[HTML]{000000} 0.2109 & {\cellcolor[HTML]{E3F399}} \color[HTML]{000000} 0.1450 & {\cellcolor[HTML]{E2F397}} \color[HTML]{000000} 0.1513 & {\cellcolor[HTML]{E5F49B}} \color[HTML]{000000} 0.1400 & {\cellcolor[HTML]{E2F397}} \color[HTML]{000000} 0.1542 & {\cellcolor[HTML]{DFF293}} \color[HTML]{000000} 0.1647 \\
\bfseries Pegasus-paraphrase & {\cellcolor[HTML]{E0F295}} \color[HTML]{000000} 0.1572 & {\cellcolor[HTML]{87CB67}} \color[HTML]{000000} 0.4963 & {\cellcolor[HTML]{ABDB6D}} \color[HTML]{000000} 0.3787 & {\cellcolor[HTML]{98D368}} \color[HTML]{000000} 0.4410 & {\cellcolor[HTML]{E0F295}} \color[HTML]{000000} 0.1570 & {\cellcolor[HTML]{A5D86A}} \color[HTML]{000000} 0.4026 & {\cellcolor[HTML]{B1DE71}} \color[HTML]{000000} 0.3532 & {\cellcolor[HTML]{89CC67}} \color[HTML]{000000} 0.4889 & {\cellcolor[HTML]{F4FAB0}} \color[HTML]{000000} 0.0597 & {\cellcolor[HTML]{F1F9AC}} \color[HTML]{000000} 0.0734 & {\cellcolor[HTML]{E5F49B}} \color[HTML]{000000} 0.1358 & {\cellcolor[HTML]{C3E67D}} \color[HTML]{000000} 0.2858 \\
\bfseries DIPPER & {\cellcolor[HTML]{CFEB85}} \color[HTML]{000000} 0.2405 & {\cellcolor[HTML]{E2F397}} \color[HTML]{000000} 0.1499 & {\cellcolor[HTML]{E0F295}} \color[HTML]{000000} 0.1585 & {\cellcolor[HTML]{E9F6A1}} \color[HTML]{000000} 0.1143 & {\cellcolor[HTML]{E9F6A1}} \color[HTML]{000000} 0.1153 & {\cellcolor[HTML]{EEF8A8}} \color[HTML]{000000} 0.0919 & {\cellcolor[HTML]{EEF8A8}} \color[HTML]{000000} 0.0910 & {\cellcolor[HTML]{EBF7A3}} \color[HTML]{000000} 0.1051 & {\cellcolor[HTML]{EEF8A8}} \color[HTML]{000000} 0.0885 & {\cellcolor[HTML]{EFF8AA}} \color[HTML]{000000} 0.0810 & {\cellcolor[HTML]{D1EC86}} \color[HTML]{000000} 0.2267 & {\cellcolor[HTML]{E5F49B}} \color[HTML]{000000} 0.1330 \\
\bfseries ChatGPT & {\cellcolor[HTML]{EFF8AA}} \color[HTML]{000000} 0.0847 & {\cellcolor[HTML]{F1F9AC}} \color[HTML]{000000} 0.0768 & {\cellcolor[HTML]{EEF8A8}} \color[HTML]{000000} 0.0865 & {\cellcolor[HTML]{E9F6A1}} \color[HTML]{000000} 0.1137 & {\cellcolor[HTML]{EBF7A3}} \color[HTML]{000000} 0.1071 & {\cellcolor[HTML]{F4FAB0}} \color[HTML]{000000} 0.0577 & {\cellcolor[HTML]{EEF8A8}} \color[HTML]{000000} 0.0919 & {\cellcolor[HTML]{F2FAAE}} \color[HTML]{000000} 0.0627 & {\cellcolor[HTML]{EFF8AA}} \color[HTML]{000000} 0.0804 & {\cellcolor[HTML]{EFF8AA}} \color[HTML]{000000} 0.0791 & {\cellcolor[HTML]{EFF8AA}} \color[HTML]{000000} 0.0826 & {\cellcolor[HTML]{EFF8AA}} \color[HTML]{000000} 0.0839 \\
\bfseries GPTZzzs & {\cellcolor[HTML]{FEFFBE}} \color[HTML]{000000} 0.0036 & {\cellcolor[HTML]{F5FBB2}} \color[HTML]{000000} 0.0540 & {\cellcolor[HTML]{FDFEBC}} \color[HTML]{000000} 0.0145 & {\cellcolor[HTML]{FBFDBA}} \color[HTML]{000000} 0.0189 & {\cellcolor[HTML]{ADDC6F}} \color[HTML]{000000} 0.3712 & {\cellcolor[HTML]{F8FCB6}} \color[HTML]{000000} 0.0369 & {\cellcolor[HTML]{FAFDB8}} \color[HTML]{000000} 0.0290 & {\cellcolor[HTML]{F8FCB6}} \color[HTML]{000000} 0.0364 & {\cellcolor[HTML]{FEFFBE}} \color[HTML]{000000} 0.0013 & {\cellcolor[HTML]{FEFFBE}} \color[HTML]{000000} 0.0005 & {\cellcolor[HTML]{FDFEBC}} \color[HTML]{000000} 0.0079 & {\cellcolor[HTML]{F5FBB2}} \color[HTML]{000000} 0.0522 \\
\bfseries GPTZeroBypasser & {\cellcolor[HTML]{AFDD70}} \color[HTML]{000000} 0.3639 & {\cellcolor[HTML]{89CC67}} \color[HTML]{000000} 0.4910 & {\cellcolor[HTML]{CBE982}} \color[HTML]{000000} 0.2516 & {\cellcolor[HTML]{66BD63}} \color[HTML]{000000} 0.5986 & {\cellcolor[HTML]{78C565}} \color[HTML]{000000} 0.5430 & {\cellcolor[HTML]{42AC5A}} \color[HTML]{000000} 0.6894 & {\cellcolor[HTML]{48AE5C}} \color[HTML]{000000} 0.6778 & {\cellcolor[HTML]{3CA959}} \color[HTML]{000000} 0.7087 & {\cellcolor[HTML]{93D168}} \color[HTML]{000000} 0.4551 & {\cellcolor[HTML]{ADDC6F}} \color[HTML]{000000} 0.3692 & {\cellcolor[HTML]{D7EE8A}} \color[HTML]{000000} 0.2067 & {\cellcolor[HTML]{89CC67}} \color[HTML]{000000} 0.4868 \\
\bfseries HomoglyphAttack & {\cellcolor[HTML]{B3DF72}} \color[HTML]{000000} 0.3446 & {\cellcolor[HTML]{36A657}} \color[HTML]{000000} 0.7220 & {\cellcolor[HTML]{A0D669}} \color[HTML]{000000} 0.4143 & {\cellcolor[HTML]{60BA62}} \color[HTML]{000000} 0.6106 & {\cellcolor[HTML]{30A356}} \color[HTML]{000000} 0.7376 & {\cellcolor[HTML]{48AE5C}} \color[HTML]{000000} 0.6785 & {\cellcolor[HTML]{30A356}} \color[HTML]{000000} 0.7383 & {\cellcolor[HTML]{33A456}} \color[HTML]{000000} 0.7286 & {\cellcolor[HTML]{8ECF67}} \color[HTML]{000000} 0.4699 & {\cellcolor[HTML]{A9DA6C}} \color[HTML]{000000} 0.3848 & {\cellcolor[HTML]{EFF8AA}} \color[HTML]{000000} 0.0858 & {\cellcolor[HTML]{7AC665}} \color[HTML]{000000} 0.5377 \\
\bfseries ALISON & {\cellcolor[HTML]{FBFDBA}} \color[HTML]{000000} 0.0201 & {\cellcolor[HTML]{FAFDB8}} \color[HTML]{000000} 0.0283 & {\cellcolor[HTML]{FBFDBA}} \color[HTML]{000000} 0.0171 & {\cellcolor[HTML]{FAFDB8}} \color[HTML]{000000} 0.0261 & {\cellcolor[HTML]{F1F9AC}} \color[HTML]{000000} 0.0739 & {\cellcolor[HTML]{FBFDBA}} \color[HTML]{000000} 0.0216 & {\cellcolor[HTML]{FAFDB8}} \color[HTML]{000000} 0.0240 & {\cellcolor[HTML]{FBFDBA}} \color[HTML]{000000} 0.0221 & {\cellcolor[HTML]{FDFEBC}} \color[HTML]{000000} 0.0154 & {\cellcolor[HTML]{FAFDB8}} \color[HTML]{000000} 0.0245 & {\cellcolor[HTML]{FEFFBE}} \color[HTML]{000000} 0.0071 & {\cellcolor[HTML]{FAFDB8}} \color[HTML]{000000} 0.0255 \\
\bfseries DFTFooler & {\cellcolor[HTML]{FEFFBE}} \color[HTML]{000000} 0.0023 & {\cellcolor[HTML]{D9EF8B}} \color[HTML]{000000} 0.1961 & {\cellcolor[HTML]{ECF7A6}} \color[HTML]{000000} 0.1001 & {\cellcolor[HTML]{CBE982}} \color[HTML]{000000} 0.2533 & {\cellcolor[HTML]{C9E881}} \color[HTML]{000000} 0.2625 & {\cellcolor[HTML]{E9F6A1}} \color[HTML]{000000} 0.1165 & {\cellcolor[HTML]{E0F295}} \color[HTML]{000000} 0.1603 & {\cellcolor[HTML]{E5F49B}} \color[HTML]{000000} 0.1349 & {\cellcolor[HTML]{F8FCB6}} \color[HTML]{000000} 0.0361 & {\cellcolor[HTML]{FBFDBA}} \color[HTML]{000000} 0.0228 & {\cellcolor[HTML]{FEFFBE}} \color[HTML]{000000} 0.0056 & {\cellcolor[HTML]{E8F59F}} \color[HTML]{000000} 0.1173 \\
\hline
\bfseries $\downarrow$ Average & {\cellcolor[HTML]{E3F399}} \color[HTML]{000000} 0.1410 & {\cellcolor[HTML]{C5E67E}} \color[HTML]{000000} 0.2739 & {\cellcolor[HTML]{DFF293}} \color[HTML]{000000} 0.1708 & {\cellcolor[HTML]{C9E881}} \color[HTML]{000000} 0.2582 & {\cellcolor[HTML]{C5E67E}} \color[HTML]{000000} 0.2789 & {\cellcolor[HTML]{CFEB85}} \color[HTML]{000000} 0.2403 & {\cellcolor[HTML]{C9E881}} \color[HTML]{000000} 0.2624 & {\cellcolor[HTML]{C9E881}} \color[HTML]{000000} 0.2639 & {\cellcolor[HTML]{E2F397}} \color[HTML]{000000} 0.1530 & {\cellcolor[HTML]{E6F59D}} \color[HTML]{000000} 0.1317 & {\cellcolor[HTML]{EBF7A3}} \color[HTML]{000000} 0.1028 &   \\
\hline
\end{tabular}
}
\caption{Ablation of attack success rate of AO methods based on classification thresholds of individual MGT detection methods optimized for each language separately by using 10\% of original (unobfuscated) test data. Only MGT detection methods achieving above 0.8 AUC ROC per each language on original test data are included.}
\label{tab:abl-asr}
\end{table*}

For ablation study regarding attack success rate (ASR) experiments, in which we included only detection methods achieving above 0.8 AUC ROC per each language (based on original MULTITuDE test data subset) and data generated by only best three LLM generators, we provide full coverage (all detection methods, all LLM generated data) in Table~\ref{tab:abl-asr-all}. Similarly, the results calculated when the classification thresholds are calibrated by using only 10\% of the test data are provided in Table~\ref{tab:abl-asr}. Although the values are slightly different from Table~\ref{tab:exp-asr}, the general conclusions hold.

\begin{table*}[!t]
\centering
\resizebox{\linewidth}{!}{
\begin{tabular}{ll|ccccccccccc}
\hline
 &  & \multicolumn{11}{c}{\bfseries Test Language [mean]} \\
 & \bfseries Detector & \bfseries ar & \bfseries ca & \bfseries cs & \bfseries de & \bfseries en & \bfseries es & \bfseries nl & \bfseries pt & \bfseries ru & \bfseries uk & \bfseries zh \\
\hline
\multirow[c]{3}{*}{\bfseries a)} & \bfseries MFD & {\cellcolor[HTML]{07753E}} \color[HTML]{F1F1F1} 0.9427 & {\cellcolor[HTML]{097940}} \color[HTML]{F1F1F1} 0.9293 & {\cellcolor[HTML]{279F53}} \color[HTML]{000000} 0.7589 & {\cellcolor[HTML]{138C4A}} \color[HTML]{F1F1F1} 0.8467 & {\cellcolor[HTML]{33A456}} \color[HTML]{000000} 0.7289 & {\cellcolor[HTML]{0E8245}} \color[HTML]{F1F1F1} 0.8888 & {\cellcolor[HTML]{148E4B}} \color[HTML]{F1F1F1} 0.8417 & {\cellcolor[HTML]{36A657}} \color[HTML]{000000} 0.7203 & {\cellcolor[HTML]{0B7D42}} \color[HTML]{F1F1F1} 0.9071 & {\cellcolor[HTML]{036E3A}} \color[HTML]{F1F1F1} 0.9699 & {\cellcolor[HTML]{BBE278}} \color[HTML]{000000} 0.3180 \\
\bfseries  & \bfseries RoBERTa-base-OpenAI-Detector & {\cellcolor[HTML]{016A38}} \color[HTML]{F1F1F1} 0.9855 & {\cellcolor[HTML]{4BB05C}} \color[HTML]{000000} 0.6718 & {\cellcolor[HTML]{06733D}} \color[HTML]{F1F1F1} 0.9501 & {\cellcolor[HTML]{0F8446}} \color[HTML]{F1F1F1} 0.8781 & {\cellcolor[HTML]{006837}} \color[HTML]{F1F1F1} 0.9941 & {\cellcolor[HTML]{C7E77F}} \color[HTML]{000000} 0.2658 & {\cellcolor[HTML]{108647}} \color[HTML]{F1F1F1} 0.8682 & {\cellcolor[HTML]{BDE379}} \color[HTML]{000000} 0.3069 & {\cellcolor[HTML]{006837}} \color[HTML]{F1F1F1} 1.0000 & {\cellcolor[HTML]{0E8245}} \color[HTML]{F1F1F1} 0.8889 & {\cellcolor[HTML]{1B9950}} \color[HTML]{F1F1F1} 0.7921 \\
\bfseries  & \bfseries XLM-RoBERTa-large (en) & {\cellcolor[HTML]{4EB15D}} \color[HTML]{000000} 0.6619 & {\cellcolor[HTML]{0B7D42}} \color[HTML]{F1F1F1} 0.9105 & {\cellcolor[HTML]{0A7B41}} \color[HTML]{F1F1F1} 0.9172 & {\cellcolor[HTML]{219C52}} \color[HTML]{F1F1F1} 0.7779 & {\cellcolor[HTML]{ABDB6D}} \color[HTML]{000000} 0.3806 & {\cellcolor[HTML]{04703B}} \color[HTML]{F1F1F1} 0.9669 & {\cellcolor[HTML]{0C7F43}} \color[HTML]{F1F1F1} 0.9025 & {\cellcolor[HTML]{026C39}} \color[HTML]{F1F1F1} 0.9828 & {\cellcolor[HTML]{06733D}} \color[HTML]{F1F1F1} 0.9530 & {\cellcolor[HTML]{036E3A}} \color[HTML]{F1F1F1} 0.9719 & {\cellcolor[HTML]{E0F295}} \color[HTML]{000000} 0.1622 \\
\hline
\multirow[c]{3}{*}{\bfseries b)} & \bfseries MFD & {\cellcolor[HTML]{FEFFBE}} \color[HTML]{000000} 0.0000 & {\cellcolor[HTML]{FEFFBE}} \color[HTML]{000000} 0.0000 & {\cellcolor[HTML]{FEFFBE}} \color[HTML]{000000} 0.0000 & {\cellcolor[HTML]{FEFFBE}} \color[HTML]{000000} 0.0000 & {\cellcolor[HTML]{FEFFBE}} \color[HTML]{000000} 0.0000 & {\cellcolor[HTML]{FEFFBE}} \color[HTML]{000000} 0.0000 & {\cellcolor[HTML]{FEFFBE}} \color[HTML]{000000} 0.0000 & {\cellcolor[HTML]{FEFFBE}} \color[HTML]{000000} 0.0000 & {\cellcolor[HTML]{FEFFBE}} \color[HTML]{000000} 0.0000 & {\cellcolor[HTML]{FEFFBE}} \color[HTML]{000000} 0.0000 & {\cellcolor[HTML]{FEFFBE}} \color[HTML]{000000} 0.0000 \\
\bfseries  & \bfseries RoBERTa-base-OpenAI-Detector & {\cellcolor[HTML]{4BB05C}} \color[HTML]{000000} 0.6662 & {\cellcolor[HTML]{D9EF8B}} \color[HTML]{000000} 0.2029 & {\cellcolor[HTML]{A7D96B}} \color[HTML]{000000} 0.3918 & {\cellcolor[HTML]{66BD63}} \color[HTML]{000000} 0.5938 & {\cellcolor[HTML]{026C39}} \color[HTML]{F1F1F1} 0.9776 & {\cellcolor[HTML]{DAF08D}} \color[HTML]{000000} 0.1907 & {\cellcolor[HTML]{78C565}} \color[HTML]{000000} 0.5462 & {\cellcolor[HTML]{C3E67D}} \color[HTML]{000000} 0.2855 & {\cellcolor[HTML]{75C465}} \color[HTML]{000000} 0.5481 & {\cellcolor[HTML]{D3EC87}} \color[HTML]{000000} 0.2253 & {\cellcolor[HTML]{5DB961}} \color[HTML]{000000} 0.6237 \\
\bfseries  & \bfseries XLM-RoBERTa-large (en) & {\cellcolor[HTML]{EFF8AA}} \color[HTML]{000000} 0.0826 & {\cellcolor[HTML]{E6F59D}} \color[HTML]{000000} 0.1299 & {\cellcolor[HTML]{E5F49B}} \color[HTML]{000000} 0.1376 & {\cellcolor[HTML]{E9F6A1}} \color[HTML]{000000} 0.1152 & {\cellcolor[HTML]{F7FCB4}} \color[HTML]{000000} 0.0454 & {\cellcolor[HTML]{DAF08D}} \color[HTML]{000000} 0.1884 & {\cellcolor[HTML]{E3F399}} \color[HTML]{000000} 0.1456 & {\cellcolor[HTML]{DCF08F}} \color[HTML]{000000} 0.1837 & {\cellcolor[HTML]{DAF08D}} \color[HTML]{000000} 0.1905 & {\cellcolor[HTML]{D1EC86}} \color[HTML]{000000} 0.2283 & {\cellcolor[HTML]{F7FCB4}} \color[HTML]{000000} 0.0468 \\
\hline
\end{tabular}
}
\caption{Ablation of attack success rate of the best-performing AO method \textit{HomoglyphAttack} for the selected MGT detection methods. a) original settings used in the experiments, b) less intensive settings.}
\label{tab:abl-asr-ao}
\end{table*}

For ablation of attack success rate of AO methods depending on the obfuscation settings, we performed \textit{HomoglyphAttack} (as the most successful attack in our study) using a less intensive settings (changing characters to their homoglyphs with a probability of 0.01 instead of 0.1 in the original experiments). The classification threshold calibration was the same as in original experiments, i.e., the thresholds calibrated per-language using the original (unobfuscated) test data. The results for the three selected MGT detection methods (one of each category with the highest attack success rates in the original experiment) are provided in Table~\ref{tab:abl-asr-ao}.
As shown, the ASR is much lower in the less intensive settings. Using these settings, the \textit{HomoglyphAttack} was not able to fool the \textit{MFD} statistical MGT detection method for any sample in any language.
The results indicate that the achieved attack success rates are highly sensitive to the used settings for obfuscation. Even low-performing AO methods, such as ALISON, can achieve higher attack success rates with proper settings adjustments. We have not experimented with settings optimization.

\begin{figure}[!b]
    \centering
    \includegraphics[width=\linewidth]{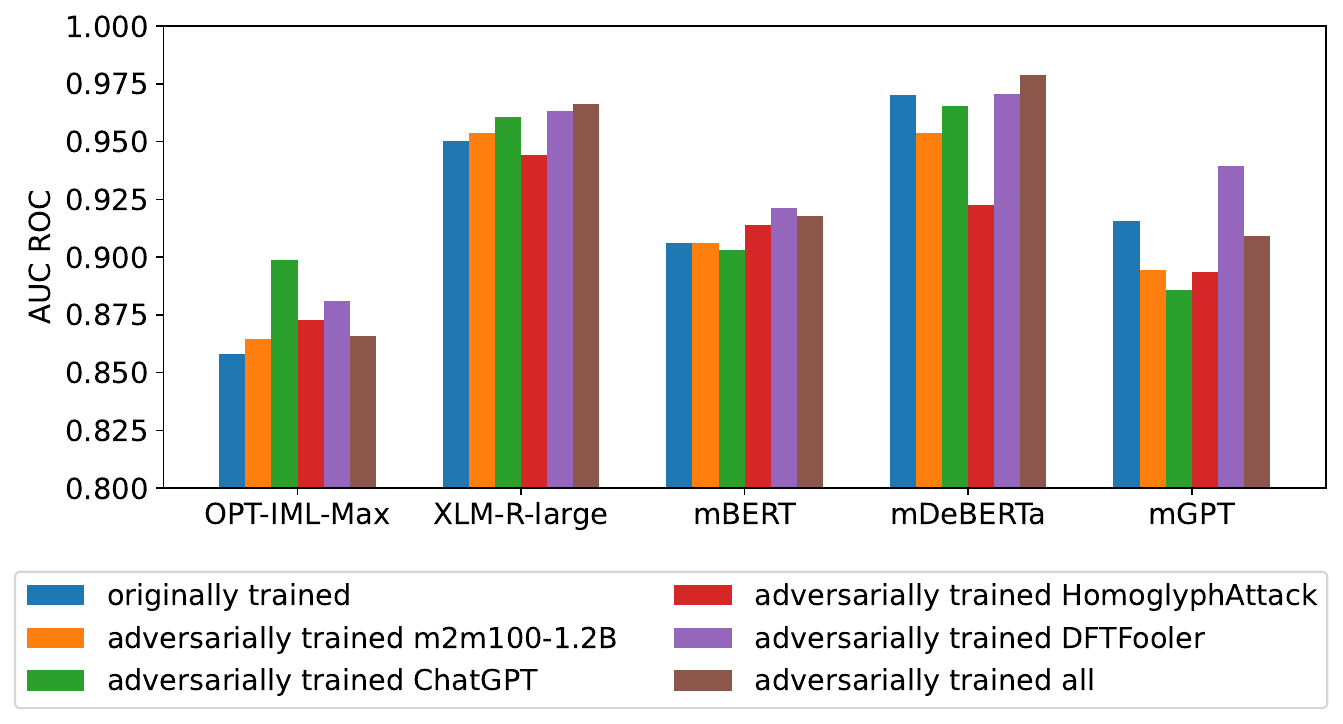}
    \caption{Ablation of detection performance (AUC ROC) of originally (leftmost bar) and adversarially trained MGT detection methods on the high-quality test data. Note that the y-axis starts at 0.8.}
    \label{fig:abl-fig-at}
\end{figure}

For ablation study regarding detection performance, such as reported in \figurename~\ref{fig:fig-at}, we provide the results in \figurename~\ref{fig:abl-fig-at} for filtered test samples, in which the obfuscated texts meet the chosen similarity metrics (reported in Table~\ref{tab:textanalysis}) thresholds in addition to not using the data of AO methods not passing the quality check in general. These thresholds include: \textit{CharLenDiff} is in the interval of $(0.5, 2)$, \textit{LD} is above 1, and \textit{METEOR, BERTScore, USE, ngram, and TF} are above $0.5$. The results show that the detection methods performance is slightly higher when using such high-quality texts than when using the whole test set including noise. The differences between originally trained and adversarially trained detection methods are lower, but shows the same trend. Nevertheless, the adversarial retraining used not only high-quality obfuscated texts, but all of them, which could affect the results.

\section{Results Data}
\label{sec:appendix-results}

\begin{table*}[!t]
\footnotesize 
\centering
\resizebox{0.95\linewidth}{!}{
\begin{tabular}{clc|ccccccccccc|c}
\hline
& \bfseries  & \bfseries  & \multicolumn{12}{c}{\bfseries Test Language [AUC ROC]} \\
\bfseries Rank &\bfseries MGT Detection Method & \bfseries Category & \bfseries ar & \bfseries ca & \bfseries cs & \bfseries de & \bfseries en & \bfseries es & \bfseries nl & \bfseries pt & \bfseries ru & \bfseries uk & \bfseries zh & \bfseries all \\
\hline
\bfseries 1 & XLM-RoBERTa-large (all) & F & {\cellcolor{lightgray}} 0.981 & {\cellcolor{lightgray}} 0.990 & {\cellcolor{lightgray}} 0.990 & {\cellcolor{lightgray}} 0.982 & {\cellcolor{lightgray}} 0.994 & {\cellcolor{lightgray}} 0.995 & {\cellcolor{lightgray}} 0.982 & {\cellcolor{lightgray}} 0.977 & {\cellcolor{lightgray}} 0.988 & {\cellcolor{lightgray}} 0.988 & {\cellcolor{lightgray}} 0.958 & {\cellcolor{lightgray}} 0.983 \\
\bfseries 2 & mDeBERTa-v3-base (all) & F & {\cellcolor{lightgray}} 0.938 & {\cellcolor{lightgray}} 0.987 & {\cellcolor{lightgray}} 0.937 & {\cellcolor{lightgray}} 0.925 & {\cellcolor{lightgray}} 0.995 & {\cellcolor{lightgray}} 0.993 & {\cellcolor{lightgray}} 0.984 & {\cellcolor{lightgray}} 0.987 & {\cellcolor{lightgray}} 0.991 & {\cellcolor{lightgray}} 0.980 & {\cellcolor{lightgray}} 0.935 & {\cellcolor{lightgray}} 0.966 \\
\bfseries 3 & XLM-RoBERTa-large (ru) & F & {\cellcolor{lightgray}} 0.964 & {\cellcolor{lightgray}} 0.967 & {\cellcolor{lightgray}} 0.995 & {\cellcolor{lightgray}} 0.981 & {\cellcolor{lightgray}} 0.959 & {\cellcolor{lightgray}} 0.983 & {\cellcolor{lightgray}} 0.971 & {\cellcolor{lightgray}} 0.966 & {\cellcolor{lightgray}} 0.993 & {\cellcolor{lightgray}} 0.981 & {\cellcolor{lightgray}} 0.934 & {\cellcolor{lightgray}} 0.954 \\
\bfseries 4 & mDeBERTa-v3-base (es) & F & {\cellcolor{lightgray}} 0.936 & {\cellcolor{lightgray}} 0.985 & {\cellcolor{lightgray}} 0.979 & {\cellcolor{lightgray}} 0.971 & {\cellcolor{lightgray}} 0.929 & {\cellcolor{lightgray}} 0.993 & {\cellcolor{lightgray}} 0.985 & {\cellcolor{lightgray}} 0.980 & {\cellcolor{lightgray}} 0.964 & {\cellcolor{lightgray}} 0.973 & {\cellcolor{lightgray}} 0.925 & {\cellcolor{lightgray}} 0.952 \\
\bfseries 5 & BERT-base-multilingual-cased (all) & F & {\cellcolor{lightgray}} 0.906 & {\cellcolor{lightgray}} 0.986 & {\cellcolor{lightgray}} 0.932 & {\cellcolor{lightgray}} 0.898 & {\cellcolor{lightgray}} 0.995 & {\cellcolor{lightgray}} 0.991 & {\cellcolor{lightgray}} 0.945 & {\cellcolor{lightgray}} 0.975 & {\cellcolor{lightgray}} 0.982 & {\cellcolor{lightgray}} 0.950 & {\cellcolor{lightgray}} 0.871 & {\cellcolor{lightgray}} 0.951 \\
\bfseries 6 & XLM-RoBERTa-large (es) & F & {\cellcolor{lightgray}} 0.965 & {\cellcolor{lightgray}} 0.971 & {\cellcolor{lightgray}} 0.984 & {\cellcolor{lightgray}} 0.974 & {\cellcolor{lightgray}} 0.839 & {\cellcolor{lightgray}} 0.986 & {\cellcolor{lightgray}} 0.979 & {\cellcolor{lightgray}} 0.968 & {\cellcolor{lightgray}} 0.971 & {\cellcolor{lightgray}} 0.972 & {\cellcolor{lightgray}} 0.957 & {\cellcolor{lightgray}} 0.946 \\
\bfseries 7 & mDeBERTa-v3-base (ru) & F & {\cellcolor{lightgray}} 0.963 & {\cellcolor{lightgray}} 0.923 & {\cellcolor{lightgray}} 0.993 & {\cellcolor{lightgray}} 0.958 & {\cellcolor{lightgray}} 0.882 & {\cellcolor{lightgray}} 0.941 & {\cellcolor{lightgray}} 0.953 & {\cellcolor{lightgray}} 0.886 & {\cellcolor{lightgray}} 0.988 & {\cellcolor{lightgray}} 0.988 & {\cellcolor{lightgray}} 0.893 & {\cellcolor{lightgray}} 0.933 \\
\bfseries 8 & BERT-base-multilingual-cased (es) & F & {\cellcolor{lightgray}} 0.877 & {\cellcolor{lightgray}} 0.980 & {\cellcolor{lightgray}} 0.938 & {\cellcolor{lightgray}} 0.940 & 0.698 & {\cellcolor{lightgray}} 0.987 & {\cellcolor{lightgray}} 0.984 & {\cellcolor{lightgray}} 0.970 & {\cellcolor{lightgray}} 0.925 & {\cellcolor{lightgray}} 0.921 & {\cellcolor{lightgray}} 0.947 & {\cellcolor{lightgray}} 0.917 \\
\bfseries 9 & mGPT (all) & F & {\cellcolor{lightgray}} 0.960 & {\cellcolor{lightgray}} 0.944 & {\cellcolor{lightgray}} 0.916 & {\cellcolor{lightgray}} 0.977 & {\cellcolor{lightgray}} 0.996 & {\cellcolor{lightgray}} 0.986 & {\cellcolor{lightgray}} 0.962 & {\cellcolor{lightgray}} 0.974 & {\cellcolor{lightgray}} 0.988 & {\cellcolor{lightgray}} 0.964 & 0.710 & {\cellcolor{lightgray}} 0.911 \\
\bfseries 10 & mGPT (ru) & F & {\cellcolor{lightgray}} 0.978 & {\cellcolor{lightgray}} 0.944 & {\cellcolor{lightgray}} 0.944 & {\cellcolor{lightgray}} 0.972 & 0.757 & {\cellcolor{lightgray}} 0.960 & {\cellcolor{lightgray}} 0.939 & {\cellcolor{lightgray}} 0.922 & {\cellcolor{lightgray}} 0.987 & {\cellcolor{lightgray}} 0.975 & 0.741 & {\cellcolor{lightgray}} 0.906 \\
\bfseries 11 & mGPT (es) & F & {\cellcolor{lightgray}} 0.930 & {\cellcolor{lightgray}} 0.970 & {\cellcolor{lightgray}} 0.901 & {\cellcolor{lightgray}} 0.976 & {\cellcolor{lightgray}} 0.939 & {\cellcolor{lightgray}} 0.991 & {\cellcolor{lightgray}} 0.989 & {\cellcolor{lightgray}} 0.980 & {\cellcolor{lightgray}} 0.964 & {\cellcolor{lightgray}} 0.957 & {\cellcolor{lightgray}} 0.904 & {\cellcolor{lightgray}} 0.906 \\
\bfseries 12 & BERT-base-multilingual-cased (ru) & F & {\cellcolor{lightgray}} 0.933 & {\cellcolor{lightgray}} 0.915 & {\cellcolor{lightgray}} 0.943 & {\cellcolor{lightgray}} 0.893 & 0.714 & {\cellcolor{lightgray}} 0.890 & {\cellcolor{lightgray}} 0.926 & {\cellcolor{lightgray}} 0.902 & {\cellcolor{lightgray}} 0.978 & {\cellcolor{lightgray}} 0.953 & {\cellcolor{lightgray}} 0.838 & {\cellcolor{lightgray}} 0.890 \\
\bfseries 13 & XLM-RoBERTa-large (en) & F & {\cellcolor{lightgray}} 0.836 & {\cellcolor{lightgray}} 0.976 & {\cellcolor{lightgray}} 0.934 & {\cellcolor{lightgray}} 0.948 & {\cellcolor{lightgray}} 0.998 & {\cellcolor{lightgray}} 0.909 & {\cellcolor{lightgray}} 0.943 & {\cellcolor{lightgray}} 0.890 & {\cellcolor{lightgray}} 0.926 & {\cellcolor{lightgray}} 0.898 & {\cellcolor{lightgray}} 0.909 & {\cellcolor{lightgray}} 0.874 \\
\bfseries 14 & OPT-IML-Max-1.3B (all) & F & 0.486 & {\cellcolor{lightgray}} 0.957 & {\cellcolor{lightgray}} 0.933 & {\cellcolor{lightgray}} 0.903 & {\cellcolor{lightgray}} 0.996 & {\cellcolor{lightgray}} 0.981 & {\cellcolor{lightgray}} 0.929 & {\cellcolor{lightgray}} 0.982 & {\cellcolor{lightgray}} 0.913 & {\cellcolor{lightgray}} 0.807 & 0.393 & {\cellcolor{lightgray}} 0.858 \\
\bfseries 15 & BERT-base-multilingual-cased (en) & F & 0.725 & {\cellcolor{lightgray}} 0.967 & {\cellcolor{lightgray}} 0.891 & {\cellcolor{lightgray}} 0.840 & {\cellcolor{lightgray}} 0.996 & {\cellcolor{lightgray}} 0.898 & {\cellcolor{lightgray}} 0.900 & {\cellcolor{lightgray}} 0.891 & {\cellcolor{lightgray}} 0.823 & {\cellcolor{lightgray}} 0.885 & 0.740 & {\cellcolor{lightgray}} 0.856 \\
\bfseries 16 & MFD & S & 0.727 & 0.766 & 0.703 & {\cellcolor{lightgray}} 0.899 & {\cellcolor{lightgray}} 0.955 & {\cellcolor{lightgray}} 0.945 & {\cellcolor{lightgray}} 0.901 & {\cellcolor{lightgray}} 0.938 & {\cellcolor{lightgray}} 0.837 & {\cellcolor{lightgray}} 0.858 & 0.761 & {\cellcolor{lightgray}} 0.833 \\
\bfseries 17 & OPT-IML-Max-1.3B (es) & F & 0.685 & {\cellcolor{lightgray}} 0.948 & {\cellcolor{lightgray}} 0.868 & {\cellcolor{lightgray}} 0.906 & {\cellcolor{lightgray}} 0.896 & {\cellcolor{lightgray}} 0.982 & {\cellcolor{lightgray}} 0.958 & {\cellcolor{lightgray}} 0.981 & 0.616 & 0.641 & 0.450 & {\cellcolor{lightgray}} 0.812 \\
\bfseries 18 & mDeBERTa-v3-base (en) & F & 0.602 & {\cellcolor{lightgray}} 0.907 & {\cellcolor{lightgray}} 0.869 & 0.765 & {\cellcolor{lightgray}} 0.998 & {\cellcolor{lightgray}} 0.938 & {\cellcolor{lightgray}} 0.856 & 0.780 & 0.703 & {\cellcolor{lightgray}} 0.835 & 0.547 & {\cellcolor{lightgray}} 0.802 \\
\bfseries 19 & DetectLLM-LRR & S & 0.659 & {\cellcolor{lightgray}} 0.931 & {\cellcolor{lightgray}} 0.886 & {\cellcolor{lightgray}} 0.881 & {\cellcolor{lightgray}} 0.939 & {\cellcolor{lightgray}} 0.911 & {\cellcolor{lightgray}} 0.938 & {\cellcolor{lightgray}} 0.887 & 0.734 & 0.764 & 0.663 & 0.791 \\
\bfseries 20 & mGPT (en) & F & 0.511 & {\cellcolor{lightgray}} 0.884 & {\cellcolor{lightgray}} 0.863 & {\cellcolor{lightgray}} 0.860 & {\cellcolor{lightgray}} 0.997 & {\cellcolor{lightgray}} 0.936 & {\cellcolor{lightgray}} 0.884 & {\cellcolor{lightgray}} 0.899 & 0.795 & 0.796 & 0.596 & 0.782 \\
\bfseries 21 & LLMDeviation & S & 0.615 & {\cellcolor{lightgray}} 0.961 & {\cellcolor{lightgray}} 0.905 & {\cellcolor{lightgray}} 0.876 & {\cellcolor{lightgray}} 0.965 & {\cellcolor{lightgray}} 0.911 & {\cellcolor{lightgray}} 0.958 & {\cellcolor{lightgray}} 0.896 & 0.684 & 0.775 & 0.683 & 0.765 \\
\bfseries 22 & GLTR Test 2 & S & 0.599 & {\cellcolor{lightgray}} 0.893 & {\cellcolor{lightgray}} 0.874 & {\cellcolor{lightgray}} 0.850 & {\cellcolor{lightgray}} 0.943 & {\cellcolor{lightgray}} 0.912 & {\cellcolor{lightgray}} 0.939 & {\cellcolor{lightgray}} 0.914 & 0.672 & 0.791 & 0.702 & 0.759 \\
\bfseries 23 & LogRank & S & 0.596 & {\cellcolor{lightgray}} 0.965 & {\cellcolor{lightgray}} 0.917 & {\cellcolor{lightgray}} 0.873 & {\cellcolor{lightgray}} 0.972 & {\cellcolor{lightgray}} 0.916 & {\cellcolor{lightgray}} 0.960 & {\cellcolor{lightgray}} 0.904 & 0.669 & 0.762 & 0.689 & 0.758 \\
\bfseries 24 & LogLikelihood & S & 0.581 & {\cellcolor{lightgray}} 0.963 & {\cellcolor{lightgray}} 0.912 & {\cellcolor{lightgray}} 0.853 & {\cellcolor{lightgray}} 0.971 & {\cellcolor{lightgray}} 0.907 & {\cellcolor{lightgray}} 0.960 & {\cellcolor{lightgray}} 0.900 & 0.634 & 0.738 & 0.685 & 0.743 \\
\bfseries 25 & OPT-IML-Max-1.3B (ru) & F & 0.518 & 0.562 & {\cellcolor{lightgray}} 0.865 & 0.725 & 0.714 & 0.742 & 0.703 & 0.704 & {\cellcolor{lightgray}} 0.918 & {\cellcolor{lightgray}} 0.815 & 0.603 & 0.696 \\
\bfseries 26 & Rank & S & 0.558 & {\cellcolor{lightgray}} 0.869 & 0.729 & 0.767 & {\cellcolor{lightgray}} 0.829 & 0.769 & {\cellcolor{lightgray}} 0.875 & 0.736 & 0.598 & 0.511 & 0.606 & 0.683 \\
\bfseries 27 & OPT-IML-Max-1.3B (en) & F & 0.353 & {\cellcolor{lightgray}} 0.832 & 0.770 & 0.788 & {\cellcolor{lightgray}} 0.997 & 0.758 & 0.663 & 0.785 & 0.480 & 0.497 & 0.373 & 0.674 \\
\bfseries 28 & ChatGPT-Detector-RoBERTa-Chinese & P & 0.734 & {\cellcolor{lightgray}} 0.819 & 0.547 & 0.709 & 0.564 & 0.536 & {\cellcolor{lightgray}} 0.854 & 0.618 & 0.615 & 0.509 & {\cellcolor{lightgray}} 0.918 & 0.638 \\
\bfseries 29 & DetectLLM-NPR & S & 0.527 & 0.758 & 0.685 & 0.628 & 0.690 & 0.687 & 0.711 & 0.689 & 0.668 & 0.717 & 0.506 & 0.636 \\
\bfseries 30 & DetectGPT & S & 0.452 & 0.765 & 0.792 & 0.584 & 0.685 & 0.630 & 0.640 & 0.652 & 0.616 & 0.635 & 0.503 & 0.593 \\
\bfseries 31 & Longformer Detector & P & 0.532 & 0.677 & 0.513 & 0.621 & {\cellcolor{lightgray}} 0.974 & 0.782 & 0.726 & 0.672 & 0.476 & 0.472 & 0.551 & 0.588 \\
\bfseries 32 & ChatGPT-Detector-RoBERTa & P & 0.513 & 0.502 & 0.427 & 0.736 & {\cellcolor{lightgray}} 0.855 & 0.709 & 0.681 & 0.640 & 0.489 & 0.461 & 0.628 & 0.570 \\
\bfseries 33 & RoBERTa-large-OpenAI-Detector & P & 0.688 & 0.479 & 0.471 & 0.448 & {\cellcolor{lightgray}} 0.927 & 0.564 & 0.412 & 0.730 & 0.614 & 0.585 & 0.475 & 0.544 \\
\bfseries 34 & RoBERTa-base-OpenAI-Detector & P & 0.637 & 0.510 & 0.574 & 0.480 & {\cellcolor{lightgray}} 0.952 & 0.518 & 0.440 & 0.504 & 0.707 & 0.455 & 0.430 & 0.542 \\
\bfseries 35 & ruRoBERTa-ruatd-binary & P & 0.577 & 0.505 & 0.501 & 0.484 & 0.611 & 0.496 & 0.507 & 0.495 & 0.619 & 0.591 & 0.505 & 0.535 \\
\bfseries 36 & RoBERTa-base-autextification-Detection & P & 0.500 & 0.478 & 0.501 & 0.509 & 0.500 & 0.533 & 0.351 & 0.512 & 0.500 & 0.500 & 0.500 & 0.489 \\
\bfseries 37 & Entropy & S & 0.563 & 0.117 & 0.164 & 0.417 & 0.303 & 0.373 & 0.151 & 0.346 & 0.584 & 0.425 & 0.539 & 0.419 \\
\hline
\end{tabular}
}
\caption{Per-language detection performance (AUC ROC) comparison of all MGT detection methods using original MULTITuDE test data (human and machine samples of the 3 selected best LLM generators -- gpt-4, gpt-3.5-turbo, vicuna). F refers to fine-tuned, P to pre-trained, and S to statistical MGT detection methods category. MGT detection methods achieving above threshold (0.8) performance are highlighted.}
\label{tab:benchmark-auc}
\end{table*}

Table~\ref{tab:benchmark-auc} shows a per-language comparison of AUC ROC detection performance of all MGT detection methods using subset of MULTITuDE test data, namely all human samples along with the machine samples generated by the selected three best LLM generators (gpt-4, gpt-3.5-turbo, vicuna), generating the texts most similar to human counterparts based on automated similarity metrics. Tables~\ref{tab:exp-auc-f}--\ref{tab:exp-auc-s} include resulted mean values (across detection methods) of AUC ROC detection performance per each AO method data and per each test language. To see differences between MGT detection method categories, we provide the results separately for them. Furthermore, we provide AUC ROC drop in Tables~\ref{tab:exp-drop-f}--\ref{tab:exp-drop-s}, reflecting the robustness of MGT detection method categories in regard to each AO method.

Table~\ref{tab:exp-drop-at-diff} provides the results of per-language comparison of AUC ROC drop caused by obfuscated data before and after adversarial retraining, reflecting whether the adversarial robustness is increased or decreased. As the results indicate, it is increased in most cases (only in four cases it was slightly decreased, even those are not statistically significant). For example, HomoglyphAttack caused mean AUC ROC drop of $-17.68\%$ for the German texts in originally trained MGT detection methods and only of $-3.21\%$ in adversarially trained methods; thus, resulting in $14.47\%$ difference.

\begin{table*}[!t]
\centering
\resizebox{0.9\linewidth}{!}{

}
\caption{AUC ROC per-language mean value along with 95\% confidence interval error bounds of finetuned MGT detection methods. A higher value indicates better detection performance.
}
\label{tab:exp-auc-f}
\end{table*}

\begin{table*}[!t]
\centering
\resizebox{0.9\linewidth}{!}{
}
\caption{AUC ROC per-language mean value along with 95\% confidence interval error bounds of pretrained MGT detection methods. A higher value indicates better detection performance.}
\label{tab:exp-auc-p}
\end{table*}

\begin{table*}[!t]
\centering
\resizebox{0.9\linewidth}{!}{
}
\caption{AUC ROC per-language mean value along with 95\% confidence interval error bounds of statistical MGT detection methods. A higher value indicates better detection performance.}
\label{tab:exp-auc-s}
\end{table*}

\begin{table*}[!t]

\centering
\resizebox{\linewidth}{!}{
}
\caption{AUC ROC drop of finetuned MGT detection methods. Per-language mean value is reported along with 95\% confidence interval error bounds. A higher drop (i.e., a lower value -- a darker color) indicates that the detection ability is affected by the corresponding AO method in a greater scale.}
\label{tab:exp-drop-f}
\end{table*}

\begin{table*}[!t]
\centering
\resizebox{\linewidth}{!}{
}
\caption{AUC ROC drop of pretrained MGT detection methods. Per-language mean value is reported along with 95\% confidence interval error bounds. A higher drop (i.e., a lower value -- a darker color) indicates that the detection ability is affected by the corresponding AO method in a greater scale.}
\label{tab:exp-drop-p}
\end{table*}

\begin{table*}[!t]
\centering
\resizebox{\linewidth}{!}{
}
\caption{AUC ROC drop of statistical MGT detection methods. Per-language mean value is reported along with 95\% confidence interval error bounds. A higher drop (i.e., a lower value -- a darker color) indicates that the detection ability is affected by the corresponding AO method in a greater scale.}
\label{tab:exp-drop-s}
\end{table*}

\begin{table*}[!t]
\centering
\resizebox{0.9\linewidth}{!}{
}
\caption{Per-language AUC ROC drop difference between originally and adversarially trained MGT detection methods. Per-language mean difference value is reported for each AO method along with 95\% confidence interval error bounds. The results that are not statistically significant are marked by (n.s.). A higher number represents more robust MGT detection after adversarial retraining.
}
\label{tab:exp-drop-at-diff}
\end{table*}

\end{document}